\pdfoutput=1

\documentclass{article}

\PassOptionsToPackage{numbers,compress}{natbib}
\usepackage[main,final]{neurips_2026}
\setcitestyle{numbers,square,comma,sort&compress}

\usepackage[utf8]{inputenc}
\usepackage[T1]{fontenc}
\usepackage{url}
\usepackage{microtype}
\usepackage{graphicx}
\usepackage[font=small,labelfont=bf,labelsep=colon,skip=5pt]{caption}
\usepackage{booktabs}
\usepackage{amsfonts}
\usepackage{nicefrac}
\usepackage{xcolor}
\usepackage{tikz}
\usetikzlibrary{arrows.meta,calc}
\usepackage[skins,breakable]{tcolorbox}
\usepackage{wrapfig}
\usepackage{array,makecell}
\usepackage{threeparttable}
\usepackage{amsmath,amssymb,amsthm,mathtools}
\usepackage{algorithm}
\usepackage[noend]{algpseudocode}
\algrenewcommand\algorithmicrequire{\textbf{Input:}}
\algrenewcommand\alglinenumber[1]{\footnotesize\textcolor{black!50}{#1}}
\algrenewcommand\algorithmiccomment[1]{\hfill\textcolor{black!55}{$\triangleright$~#1}}
\newcommand{\AlgPhase}[1]{\item[\textcolor{black!65}{\itshape #1}]}
\usepackage{enumitem}
\usepackage{afterpage}
\usepackage{placeins}
\newcommand{\Needspace}[1]{\par\begingroup\dimen0=#1\relax\vskip0pt plus\dimen0\penalty-100\vskip0pt plus-\dimen0\vskip\dimen0\penalty9999\vskip-\dimen0\vskip0pt\endgroup}
\usepackage{pifont}
\usepackage{etoc}
\usepackage{hyperref}
\definecolor{linkblue}{RGB}{25,80,150}
\hypersetup{colorlinks=true,linkcolor=linkblue,citecolor=linkblue,urlcolor=linkblue,linktoc=page,
  pdftitle={TRAM: Test-Time Risk Adaptation with Mixture of Agents},
  pdfauthor={Mohamad Chehade, Amrit Singh Bedi, Souradip Chakraborty, Amy Zhang, Hao Zhu},
  pdfsubject={NeurIPS 2026},
  pdfkeywords={reinforcement learning, risk adaptation, transfer learning}}

\usepackage{amsmath,amsfonts,bm}

\def\eqref#1{\ref{#1}}

\def\1{\bm{1}}

\DeclareMathAlphabet{\mathsfit}{\encodingdefault}{\sfdefault}{m}{sl}
\SetMathAlphabet{\mathsfit}{bold}{\encodingdefault}{\sfdefault}{bx}{n}

\newif\ifshowchanges
\showchangesfalse

\newcommand{\Cmark}{\ding{51}}
\newcommand{\Xmark}{\ding{55}}
\newcolumntype{C}[1]{>{\centering\arraybackslash}p{#1}}
\newcolumntype{L}[1]{>{\raggedright\arraybackslash}p{#1}}

\allowdisplaybreaks
\tcbset{
  tramBox/.style={enhanced,unbreakable,boxrule=0.5pt,arc=1.2mm,left=4pt,right=4pt,top=4pt,bottom=4pt,before skip=5pt,after skip=5pt},
  theoremBox/.style={tramBox,colback=blue!3!white,colframe=blue!45!black},
  supportBox/.style={tramBox,colback=gray!3!white,colframe=gray!50!white,arc=1.5pt},
  equationBox/.style={tramBox,colback=resultblue!3!white,colframe=resultblue!35!white,arc=1.5pt},
  insightBox/.style={tramBox,colback=green!4!white,colframe=green!45!black}
}
\newtcolorbox{goalbox}{supportBox}
\newtcolorbox{assumptionbox}{supportBox}
\newtcolorbox{stepbox}{supportBox}
\newtcolorbox{keybox}{equationBox}
\newtcolorbox{remarkbox}{supportBox}
\newtcolorbox{insightbox}{insightBox}

\numberwithin{equation}{section}
\theoremstyle{plain}
\newtheorem{theorem}{Theorem}[section]
\newtheorem{lemma}[theorem]{Lemma}
\newtheorem{proposition}[theorem]{Proposition}
\newtheorem{corollary}[theorem]{Corollary}
\theoremstyle{definition}
\newtheorem{definition}[theorem]{Definition}
\theoremstyle{remark}

\definecolor{resultblue}{RGB}{0,84,180}
\tcbset{
  stmtBox/.style={enhanced,unbreakable,boxrule=0.5pt,arc=1.5pt,left=5pt,right=5pt,top=3pt,bottom=3pt,
                  before skip=6pt plus 2pt,after skip=6pt plus 2pt,
                  before upper={\setlength{\topsep}{0pt}\setlength{\partopsep}{0pt}}},
  resultStmt/.style={stmtBox,colback=resultblue!4!white,colframe=resultblue!70!black},
  supportStmt/.style={stmtBox,colback=gray!3!white,colframe=gray!50!white}
}
\tcolorboxenvironment{theorem}{resultStmt}
\tcolorboxenvironment{proposition}{resultStmt}
\tcolorboxenvironment{corollary}{resultStmt}
\tcolorboxenvironment{lemma}{supportStmt}
\tcolorboxenvironment{definition}{supportStmt}

\title{TRAM: Test-Time Risk Adaptation\\with Mixture of Agents}

\author{%
Mohamad Chehade\\
{\normalfont UT Austin}\\
{\normalfont\ttfamily\small chehade@utexas.edu}
\And
Amrit Singh Bedi\\
{\normalfont University of Central Florida}\\
{\normalfont\ttfamily\small amritbedi@ucf.edu}
\And
Souradip Chakraborty\\
{\normalfont MIT}\\
{\normalfont\ttfamily\small soura24@mit.edu}
\AND
Amy Zhang\\
{\normalfont UT Austin}\\
{\normalfont\ttfamily\small amy.zhang@austin.utexas.edu}
\And
Hao Zhu\\
{\normalfont UT Austin}\\
{\normalfont\ttfamily\small haozhu@utexas.edu}
}

\begin{document}
\maketitle

\begin{abstract}
Deployed reinforcement learning agents often face safety requirements that are specified only after training: new hazard maps, revised risk thresholds, or behavioral alignment constraints. We study zero-update deployment-time adaptation in which a fixed library of risk-neutral source policies must be reused under a newly specified reward--risk tradeoff. We propose \textbf{TRAM} (\textbf{T}est-time \textbf{R}isk \textbf{A}daptation with \textbf{M}ixture of \textbf{A}gents), a source-scored composition rule that evaluates each source under the target reward and an occupancy-based deployment risk, then selects actions using risk-adjusted source scores. Unlike training-time risk-sensitive methods tied to a fixed surrogate such as return variance, TRAM supports spatial barrier exposure, divergence to a reference behavior, and local volatility risks specified at test time. We make the surrogate nature of the method explicit: TRAM is not claimed to solve the full occupancy-control problem of the stitched policy, but it admits a measurable source-hull mismatch term connecting source-scored risk to realized risk. Experiments in gridworlds, MuJoCo Reacher, Safety-Gymnasium, and an inference-time LLM alignment setting show that TRAM improves deployment risk while preserving reward and requires no parameter updates at test time.
\end{abstract}

\etocdepthtag.toc{mainmatter}
\section{Introduction}
\label{sec:introduction}
Reinforcement learning systems are often deployed after training into settings where the safety specification changes. A robot may receive a new goal or temporary keep-out region, a safe-navigation agent may face a revised hazard map, or an assistant may be judged against a new behavioral reference. These changes are usually operational rather than architectural: the already-trained behaviors are still useful, but the desired reward--risk tradeoff is not known until deployment. Throughout, a \emph{source policy} is a fixed, already-trained policy, and the \emph{source library} is the set of such policies, trained once before deployment (\emph{source training}). At deployment our method trains nothing: at each step it scores every source by its estimated value for the new task minus the requested risk penalty and acts with the best one. Retraining for each new specification is expensive, slow, and sometimes not allowed once a model is deployed. The problem is increasingly pressing as pretrained policy libraries and behavior foundation models are reused across tasks~\cite{pirotta2024fast,park2024hilp,tirinzoni2025motivo}, and as alignment of large language models shifts from retraining toward inference-time control~\cite{controlled_decoding,Souradip,snell2025scaling}. The core idea of this paper is to move the choice of risk from source training to deployment-time decision making: train reusable behaviors first, then compose them according to the risk requested at test time.

\begin{figure*}[t]
    \centering
    \includegraphics[width=\textwidth]{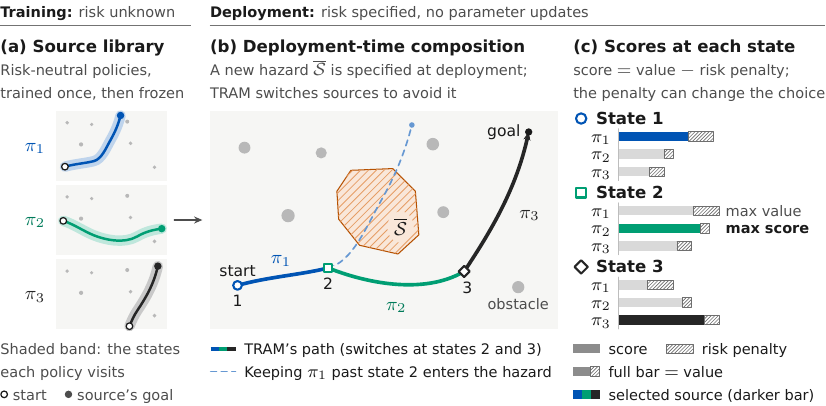}
    \caption{\textbf{TRAM decouples source training from deployment-time risk specification.}
    (a)~Before deployment, a library of risk-neutral source policies is trained once and frozen; the shaded band around each route marks the states that policy visits.
    (b)~At deployment, a new hazard $\overline{\mathcal S}$ is specified. At each state, TRAM scores every source and acts with the best one, so the executed path is stitched from the source routes in~(a); keeping $\pi_1$ past state~2 (dashed) would enter the hazard.
    (c)~A source's score is its value $\max_a Q_T^{\pi_j}(s,a)$ (full bar) minus a fixed per-source risk penalty (hatched), Eq.~\eqref{eq:source_score}. At state~2, $\pi_1$ has the highest value but $\pi_2$ the highest score, so the penalty changes the choice. TRAM selects the maximum by Eq.~\eqref{eq:TRAM_policy} and re-scores at the next state; all quantities are schematic, and no parameters are updated.}
    \label{fig:tram_local_source_selection_overview}
\end{figure*}

\paragraph{Why fixed training-time risk is brittle.}
Risk-sensitive RL typically chooses the risk model during training, such as variance, CVaR, or a fixed constraint. This is effective when the future risk is known, but brittle when the deployment risk changes from return variance to a spatial barrier, a behavioral reference, or a local volatility signal. A risk model chosen too early can over-specialize the source policies, remove useful behaviors, and leave no mechanism to recover when the deployment specification changes. Figure~\ref{fig:tram_local_source_selection_overview} summarizes the alternative: keep a reusable source library, then inject the requested risk only at deployment.

\paragraph{Proposed approach.}
We propose \textbf{TRAM} (\textbf{T}est-time \textbf{R}isk \textbf{A}daptation with \textbf{M}ixture of \textbf{A}gents), a zero-update method for adapting a fixed source library to newly specified deployment risks. TRAM trains source policies risk-neutrally to preserve behavioral coverage. At deployment, TRAM scores each candidate source action by combining target-task value with the requested occupancy-risk penalty, then selects the best source-action pair at the current state. This converts risk adaptation into fast source scoring rather than retraining, constrained RL, or source-policy fine-tuning. As shown in Figure~\ref{fig:tram_local_source_selection_overview}, the active source can change along the trajectory, producing a locally stitched behavior that responds to the current state and risk specification. TRAM is therefore a tractable \emph{source-scored surrogate}: it does not claim to exactly solve the full occupancy-control problem of the stitched policy. The main text focuses on the implementable scoring rule and its localized transfer guarantees (Theorems~\ref{thm:main_localized} and~\ref{thm:library_design}), while Appendix~\ref{appendix:stitching_limitation} analyzes the measurable mismatch between cached source risks and the realized stitched occupancy.

\paragraph{Contributions.}
\textbf{Problem:} we formalize zero-update deployment-time adaptation to newly specified occupancy risks.
\textbf{Method:} we introduce TRAM, a tractable source-scored composition rule supporting barrier, divergence, and local-volatility risks at test time.
\textbf{Theory:} we provide localized transfer bounds, a conditional source-hull comparison, and extensions for stitched occupancies and approximate evaluation.
\textbf{Evidence:} we demonstrate TRAM across gridworlds, MuJoCo Reacher, Safety-Gymnasium, and LLM alignment without deployment-time parameter updates, including a comparison with a constrained policy retrained on the task.

\section{Related Work}
\label{related_works}
\paragraph{Zero-shot transfer and successor features.}
Zero-shot RL learns representations or policies that generalize without fine-tuning~\cite{zero_shot_dynamics,zero_shot_instructions,zero_shot_darla,zero_shot_hypernetworks,zero_shot_FB}. Successor features (SFs) exploit shared structure to evaluate policies under new rewards by dot products~\cite{sf_1,sf_2,sf_3}, and can be combined with learned task-specific features~\cite{carvalho2023sfk}. Behavior foundation models scale this idea to pretrained policy families that are prompted with a reward or a few demonstrations at test time~\cite{pirotta2024fast,park2024hilp,tirinzoni2025motivo}. These methods support reward transfer, but they generally assume the evaluation objective is fixed once the target reward is known. TRAM keeps the zero-update benefit while adding a deployment-time risk term to the source score.

\paragraph{Risk-sensitive RL and safe adaptation.}
Risk-sensitive RL optimizes objectives involving variance, CVaR, constrained returns, or dual variables~\cite{risk-aware1,risk-aware2,risk-aware3,risk-aware4,risk-aware5,risk-aware6,risk-aware7,risk-aware8,risk-aware9,Amrit}. Safety-transfer methods learn teachers, safety critics, reusable safe policies, or risk-sensitive options~\cite{risk-aware_transfer_RL_alekh_agarwal,risk-aware_transfer_safety_critic,risk-aware_transfer_robot_safety,risk-aware_transfer_risk_function,risk-aware_transfer_HRL,risk-aware_transfer_option_critic,risk-aware_transfer_option_model_uncertainty}. Constraint-conditioned and dual actor--critic approaches train over constraint families or adapt dual variables after the constraint is known~\cite{yao2023constraint_conditioned,lee2023dual_variable_actor_critic}, and offline safe RL methods adjust to new cost thresholds at deployment through cost-conditioned sequence models or feasibility-guided diffusion policies~\cite{liu2023cdt,zheng2024fisor}. Distributionally robust RL hedges against dynamics or reward shifts within an ambiguity set fixed before training~\cite{iyengar2005robust,nilim2005robust,zhou2021finite,panaganti2022sample}, whereas TRAM adapts to a risk specified at deployment under fixed dynamics. These lines are complementary, but they usually require the relevant risk family during training or nontrivial optimization during adaptation.

\paragraph{Inference-time alignment and mixtures of agents.}
For language models, alignment is increasingly performed at inference time: controlled and reward-guided decoding steer a frozen model with a value or reward estimate~\cite{controlled_decoding,khanov2024args,Souradip}, and mixture-of-agents methods combine several frozen models~\cite{wang2025moa,Collab}. Weight interpolation and constrained fine-tuning address multiple or safety-critical objectives at training time~\cite{rame2023rewarded,dai2024saferlhf}, while optimizing a learned reward too strongly degrades true quality~\cite{gao2023overoptimization}. TRAM brings this inference-time, mixture-of-agents view to sequential decision-making, with an explicit risk term specified at deployment.

\paragraph{Risk-aware SFs and TRAM's positioning.}
Risk-aware successor features (RaSF)~\cite{Gimelfarb} combine SF transfer with a mean--variance risk proxy. TRAM differs in two ways: source policies are trained risk-neutrally to preserve reusable behavioral coverage, and the deployment risk is an occupancy-based functional supplied at test time, such as barrier exposure, KL-to-reference, or local volatility. Thus TRAM sits between zero-shot transfer and risk-aware decision-making: fixed sources are reused like SF/GPI, but action selection is risk-adjusted at inference by TRAM.

\begin{table*}[t]
\centering
\normalsize
\setlength{\tabcolsep}{6pt}
\renewcommand{\arraystretch}{1.2}
\begin{threeparttable}
\caption{\textbf{Comparison with representative families.} Columns: risk sensitivity, support for risk types beyond variance, cross-task transfer, use of shared task structure, and little/no test-time compute.}
\label{tab:relevant_work_tram}

{%
\begin{tabular}{@{} C{0.24\textwidth} *{5}{C{0.12\textwidth}} @{}}
\toprule
\textsc{Method} &
\makecell{\textsc{Risk-}\\\textsc{aware}} &
\makecell{\textsc{General}\\\textsc{Risk}} &
\makecell{\textsc{Transfer}} &
\makecell{\textsc{Task}\\\textsc{Struct.}} &
\makecell{\textsc{Test-}\\\textsc{time}} \\
\midrule
Risk-sensitive RL     & \Cmark & \Xmark & \Xmark & \Xmark & \Xmark \\
Zero-shot RL                             & \Xmark & \Xmark & \Cmark & \Xmark & \Cmark \\
Dual RL               & \Cmark & \Cmark & \Xmark & \Xmark & \Xmark \\
Risk-aware Adaptation                    & \Cmark & \Xmark & \Cmark & \Xmark & \Xmark \\
\textbf{TRAM (ours)}  & \Cmark & \Cmark & \Cmark & \Cmark & \Cmark \\
\bottomrule
\end{tabular}%
}

\begin{tablenotes}[flushleft]
\footnotesize
\item \textbf{Refs.} Risk-sensitive RL~\cite{risk-aware1,risk-aware2,risk-aware3,risk-aware4,risk-aware5,risk-aware6,risk-aware7,risk-aware8,risk-aware9}; zero-shot RL~\cite{zero_shot_dynamics,zero_shot_instructions,zero_shot_darla,zero_shot_hypernetworks,zero_shot_FB}; dual RL~\cite{Amrit}; risk-aware adaptation/SFs~\cite{Gimelfarb,risk-aware_transfer_RL_alekh_agarwal,risk-aware_transfer_safety_critic,risk-aware_transfer_robot_safety,risk-aware_transfer_risk_function,risk-aware_transfer_HRL,risk-aware_transfer_option_critic,risk-aware_transfer_option_model_uncertainty,sf_1,sf_2,sf_3}.
\end{tablenotes}
\end{threeparttable}
\end{table*}

\section{Problem Formulation}
\label{sec:preliminaries}
\subsection{Deployment Setting}
\label{sec:problem_setup}

\textbf{Deployment-time safety as constrained optimization.}
We model agent-environment interaction as an MDP $M=(\mathcal{S}, \mathcal{A}, p, R, \gamma)$ with state space $\mathcal{S}$, action space $\mathcal{A}$, transition kernel $p(\cdot \mid s,a)$, reward $R(s,a,s')$, and discount $\gamma\in[0,1)$. A policy $\pi$ induces action-value function $Q^\pi(s,a) = \mathbb{E}^\pi[G_t \mid S_t=s, A_t=a]$, where $G_t=\sum_{i=0}^\infty \gamma^i R_{t+i+1}$.

At deployment, the agent faces a target task $M_{\mathrm{T}}=(\mathcal{S},\mathcal{A},p_{\mathrm T},R_{\mathrm{T}},\gamma)$ with performance measured by expected return $J(\pi) = \mathbb{E}_\pi[\sum_{t\ge 0}\gamma^t R_{\mathrm{T}}(S_t,A_t,S_{t+1})]$. Deployment also introduces a safety or alignment requirement specified only at test time through a risk functional $\rho(\pi)$:
\begin{equation}
\label{eq:constrained_rl}
\max_{\pi} J(\pi) \quad \text{subject to} \quad \rho(\pi) \leq \delta,
\end{equation}
where $\delta>0$ is the risk tolerance. Since solving a new constrained RL problem at deployment can be too slow, we work with the penalized surrogate
\begin{equation}
\label{eq:penalized_surrogate}
\max_{\pi} \tilde{J}_c(\pi) = J(\pi) - c \, \rho(\pi), \quad c \geq 0.
\end{equation}

\textbf{Transfer setting: no retraining allowed.}
We assume access to fixed source policies $\{\pi_j\}_{j=1}^n$ trained on related source tasks $M_j=(\mathcal{S},\mathcal{A},p_j,R_j,\gamma)$. The finite-MDP analysis evaluates all policies under the target dynamics and assumes each source policy is optimal for its own source task under those dynamics. Changes to goals or hazard specifications can leave dynamics unchanged. When source values or occupancies are estimated using an approximate model, Appendix~\ref{app:dynamics_mismatch} accounts for the evaluation error. The token-level LLM experiment illustrates the scoring principle and is outside the occupancy-MDP guarantees.

\subsection{Why Variance-Based Risk Fails at Deployment}
\label{sec:limitations_variance}

Many risk-aware RL and transfer methods optimize a risk proxy selected before deployment, commonly return variance, as in RaSF~\cite{Gimelfarb}. These methods are valuable when the risk specification is known in advance, but they expose three limitations when the risk is revealed or changed only at deployment.

\begin{figure}[t]
    \centering
    \includegraphics[width=\textwidth]{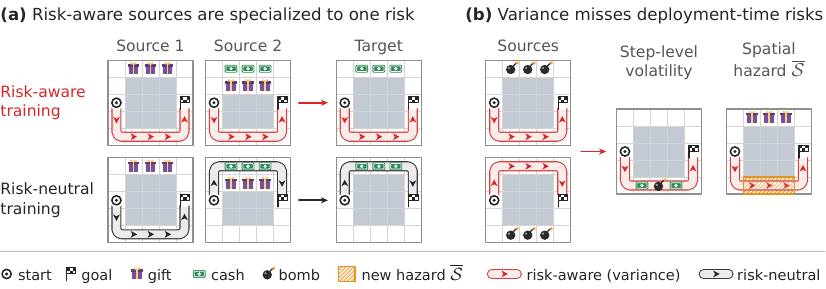}
    \caption{
    \textbf{Motivating limitations of variance-based risk-aware transfer.}
    \textbf{(a)} A risk-aware source is optimal for its training-time (variance) risk but specialized to it; when the deployment risk differs, it can miss the target-optimal route, while risk-neutral training preserves diverse reusable behaviors.
    \textbf{(b)} Trajectory-return variance can miss deployment-time risks: stochastic per-step rewards, spatial danger regions, and deployment-specific hazards are not reliably captured by return variance alone.
    }
    \label{fig:main_motivation}
\end{figure}

\textbf{Limitation 1: Conservative training destroys behavioral diversity.}
Variance-penalized or constraint-specialized sources can converge to uniformly cautious behaviors (Figure~\ref{fig:main_motivation}(a)), shrinking the repertoire available for test-time composition. Training sources risk-neutrally on varied tasks can retain useful behaviors, although diversity and source-hull inclusion are not automatic consequences of removing the training penalty.

\textbf{Limitation 2: Variance degenerates in deterministic settings.}
With deterministic dynamics and policies, trajectory return variance can be zero (Figure~\ref{fig:main_motivation}(b), center), giving no safety signal even when step-level hazards exist. Variance-based selection can therefore reduce to risk-agnostic behavior precisely when predictable environments accumulate clear step-level risks.

\textbf{Limitation 3: Variance misrepresents domain-specific risks.}
Return variance captures uncertainty in cumulative reward, not spatial exposure, behavioral drift, or worst-case contact with a new hazard (Figure~\ref{fig:main_motivation}(b), right). A low-variance path can cross a dangerous region, while a safer path may have higher benign stochasticity.

These limitations suggest three requirements for deployment-time adaptation: preserve a broad source library before the risk is known, accept risk functionals beyond return variance, and select actions without retraining or re-solving a new constrained MDP. They also motivate reporting reward, requested risk, and source coverage separately rather than collapsing all outcomes into one safe-RL score. TRAM is designed around exactly these requirements.

\section{TRAM: Test-Time Risk Adaptation via Source-Scored Composition}
\label{sec:solution_methodology}
The limitations in Section~\ref{sec:limitations_variance} motivate a deployment-time method that separates source training from the risk model revealed at deployment. TRAM aims to preserve source diversity through risk-neutral training and introduces risk only through source scoring at test time.

\paragraph{Occupancy-based deployment risks.}
Deployment risk is often not just a property of the final return. In navigation, risk may be time spent near a barrier; in robotics, contact with a forbidden region; in power-system control, repeated operation near voltage or line-flow limits. These risks depend on where the policy spends discounted probability mass, not only on the variance of its cumulative reward. Using a fixed initial distribution $\mu_0$ and target dynamics, TRAM therefore represents deployment risks as functionals of normalized discounted state-action occupancies $d^\pi$:
\begin{equation}
 d^{\pi_j}(s,a)=(1-\gamma)\sum_{t\ge0}\gamma^t
 \Pr_{\pi_j}(S_t=s,A_t=a\mid S_0\sim\mu_0).
\end{equation}
In implementation, global source occupancies are cached for the target dynamics, while values are evaluated by successor features, rollouts, or learned critics. This supports risks that return variance cannot encode: \textbf{spatial barrier exposure}, e.g.,
\begin{equation}
\rho_{\mathrm{barrier}}(d)=-\log(\tau-d(\overline{\mathcal S})),
\quad d(\overline{\mathcal S})=\sum_{s,a}d(s,a)\mathbf1\{s\in\overline{\mathcal S}\},
\end{equation}
on its domain $d(\overline{\mathcal S})<\tau$; \textbf{local volatility}, which measures per-step fluctuations under the occupancy; and \textbf{behavioral drift}, e.g., $\rho_{\mathrm{KL}}(d)=\mathrm{KL}(d\|\bar d)$ to remain close to a safe reference behavior. These functionals are specified after source training and enter only through the deployment-time score. The Lipschitz guarantees below require the compared occupancies to remain in a domain with bounded risk variation: in particular, a log barrier must stay a positive distance from its singularity. Appendix~\ref{sec:risk_functionals} states the relevant conditions.

\paragraph{Source-scored policy construction.}
Given fixed source policies $\{\pi_j\}_{j=1}^n$, TRAM computes for each source a target-task value estimate $Q_T^{\pi_j}(s,a)$ and a cached source-level risk $\rho(d^{\pi_j})$. The source-adjusted score and deployed action are the two core equations of TRAM:
\begin{align}
\label{eq:source_score}
S_j(s,a)&=Q_T^{\pi_j}(s,a)-c\rho(d^{\pi_j}),\qquad c\ge0,\\
\label{eq:TRAM_policy}
\pi_T(s)&\in\arg\max_{a\in\mathcal A}\max_{j\in[n]}S_j(s,a).
\end{align}
Algorithm~\ref{alg:TRAM} implements Eq.~\eqref{eq:TRAM_policy}: cache source occupancies and risks, evaluate source values at the current state, and select the best risk-adjusted source action. The rule is deliberately a \emph{tractable source-scored surrogate}; it reuses fixed source policies and cached source risks rather than repeatedly solving a new constrained RL problem. Appendix~\ref{appendix:stitching_limitation} gives a diagnostic for when the realized stitched occupancy differs from the source-hull proxy. An optional source screen does not certify the stitched policy's risk, because switching among eligible sources can change its occupancy.

This makes TRAM a modular deployment layer. Source training builds a library intended to provide broad behavioral coverage; deployment supplies the target reward, risk functional, tradeoff weight, and optional threshold. In new hazard maps, updated operating limits, or revised behavioral references, Algorithm~\ref{alg:TRAM} changes only the score used to select among fixed sources. The method is therefore most appropriate when source policies are already available, test-time updates are undesirable, and the requested risk can be evaluated or approximated per source.

\begin{algorithm}[t]
\caption{TRAM: zero-update source-scored risk adaptation}
\label{alg:TRAM}
\begin{algorithmic}[1]
\Require fixed sources $\{\pi_j\}_{j=1}^n$; target value specification; deployment risk $\rho$; weight $c\ge0$; optional threshold $\delta$
\AlgPhase{Before deployment: once per risk specification}
\State Estimate source occupancies under the target dynamics; cache finite risks $\widehat\rho_j$
\State $\mathcal J\gets[n]$, or $\mathcal J\gets\{j:\widehat\rho_j\le\delta\}$ if screening is used \Comment{optional source screen}
\State \algorithmicif\ $\mathcal J=\varnothing$ \algorithmicthen\ stop \Comment{the screen admits no deployment policy}
\AlgPhase{At deployment: no source parameters are updated}
\For{each deployment state $s$}
    \For{$j\in\mathcal J$ and $a\in\mathcal A$}
        \State Evaluate $\widehat Q_T^{\pi_j}(s,a)$ by successor features, a critic, or rollouts
        \State $S_j(s,a)\gets\widehat Q_T^{\pi_j}(s,a)-c\,\widehat\rho_j$ \Comment{score, Eq.~\eqref{eq:source_score}}
    \EndFor
    \State Execute $a^\star\in\arg\max_{a\in\mathcal A}\max_{j\in\mathcal J}S_j(s,a)$ \Comment{select, Eq.~\eqref{eq:TRAM_policy}}
\EndFor
\end{algorithmic}
\end{algorithm}

Appendix~\ref{app:sg_sensitivity} discusses how to choose $c$ and its relation to the tolerance $\delta$.

\paragraph{Online computation.}
For linear successor features $Q_T^{\pi_j}(s,a)=\psi^{\pi_j}(s,a)^\top w_T$, online scoring costs $\mathcal O(n|\mathcal A|d)$ per state for $n$ sources, $|\mathcal A|$ actions, and feature dimension $d$; if value tables are cached, it costs $\mathcal O(n|\mathcal A|)$. The risk term is a constant-time lookup after source occupancy caching. Thus the deployment-time work in TRAM is source scoring, not retraining, target-task policy optimization, or repeated global occupancy recomputation for every candidate stitched action. This complexity excludes source evaluation and occupancy estimation before scoring; a changed transition model can require new occupancy estimates even though source parameters remain fixed.

\subsection{When and Why TRAM Works: Theoretical Foundations}
\label{sec:theory}

We next give a compact theoretical account of the source-scored surrogate optimized by TRAM. The results isolate reward transfer quality, the deployment-risk penalty, and source-library coverage. The realized stitched-policy correction is handled separately in Appendix~\ref{appendix:stitching_limitation}, and the approximate-dynamics extension is given in Appendix~\ref{app:dynamics_mismatch}.

Traditional transfer learning bounds can require global reward similarity across all state-action pairs, often yielding pessimistic guarantees. TRAM enables a localized analysis through trajectory-specific mismatch terms. The analysis uses finite discounted MDPs, bounded rewards, common dynamics, stationary policies, and a nonempty retained source library. Write $\pi^*$ for a statewise risk-neutral target optimum. We assume each source policy $\pi_j$ is statewise optimal for its source task under the common dynamics, i.e., $Q_j^{\pi_j}=Q_j^*$. For expected one-step rewards $r_T,r_j$, define the localized reward discrepancy:
\begin{equation}
\label{eq:localized-discrepancy}
\Delta_\pi^{(T,j)}(s,a)=
\mathbb E_\pi\!\left[
 \sum_{t\ge0}\gamma^t|r_T(S_t,A_t)-r_j(S_t,A_t)|
 \,\middle|\,S_0=s,A_0=a\right].
\end{equation}
The transfer term contains discrepancies along both policies:
\[
C_j(s,a)=\Delta_{\pi^*}^{(T,j)}(s,a)
       +\Delta_{\pi_j}^{(T,j)}(s,a).
\]

\begin{theorem}[Localized risk-adjusted guarantee]
\label{thm:main_localized}
Assume exact source values and risks, and let $\rho$ be finite and $L$-Lipschitz on a domain containing all source occupancies, $d^{\pi^*}$, and $d^{\pi_T}$. Then
\begin{equation}
\widetilde Q_T^{\pi^*}(s,a)-\widetilde Q_T^{\pi_T}(s,a)
 \le\min_j C_j(s,a)+4Lc,
\end{equation}
where $\widetilde Q_T^\pi=Q_T^\pi-c\rho(d^\pi)$ uses the actual global policy occupancy.
\end{theorem}

\begin{corollary}[Reward-only guarantee]
\label{cor:standard_performance}
Under the same assumptions,
\begin{equation}
Q_T^*(s,a)-Q_T^{\pi_T}(s,a)
 \le\min_j C_j(s,a)+2Lc.
\end{equation}
\end{corollary}

\paragraph{Proof mechanism and scope.}
Set $\rho_{\min}=\min_j\rho(d^{\pi_j})$ and shift every score by $c\rho_{\min}$. This leaves the action rule unchanged and makes the source penalties nonnegative. Bellman monotonicity then gives
$Q_T^{\pi_T}\ge\max_j\{Q_T^{\pi_j}-c(\rho(d^{\pi_j})-\rho_{\min})\}$.
Comparing each source with $\pi^*$ yields the reward bound; comparing the two actual occupancy risks adds $2Lc$. Appendix~\ref{app:proofs} gives the proofs. The comparator is risk-neutral, and neither guarantee asserts hard-constraint satisfaction or risk-optimality. Uniform source-score error $\eta$ adds $2\gamma\eta/(1-\gamma)$ to both bounds (Appendix~\ref{app:score_error}). The convex-risk source-hull bridge in Appendix~\ref{app:stitching_diagnostics} provides a separate diagnostic of realized risk.

\subsection{Why Risk-Neutral Sources Are Useful}
\label{sec:risk_neutral_sources}

A central design claim in TRAM is that source policies should remain risk-neutral when the future deployment risk is unknown. Intuitively, risk-aware source training can prune away behaviors that look unsafe under one guessed risk but are valuable under a different future risk. The next result formalizes a conditional source-library design principle: broader coverage helps when the risk-neutral library's occupancy hull contains the hulls of the alternatives. Risk-neutral training alone does not guarantee that inclusion.

\begin{theorem}[Conditional source-hull dominance]
\label{thm:library_design}
Let $S_\lambda\subseteq\mathcal D$ be source occupancy libraries, with $\lambda=0$ denoting risk-neutral training. Assume
$\operatorname{co}(S_\lambda)\subseteq\operatorname{co}(S_0)$ for all $\lambda\ge0$, and that every hull admits an occupancy with $\langle d,r_T\rangle\ge V$. For a nonempty class $\mathcal R_L$ of finite $L$-Lipschitz risks on $\mathcal D$,
\begin{equation}
\sup_{\rho\in\mathcal R_L}\mathcal E_{\rho,V}(S_0)
=\inf_{\lambda\ge0}\sup_{\rho\in\mathcal R_L}\mathcal E_{\rho,V}(S_\lambda),
\end{equation}
where the nonnegative risk regret is
\[
\mathcal E_{\rho,V}(S)=
 \inf_{\substack{d\in\operatorname{co}(S)\\\langle d,r_T\rangle\ge V}}\rho(d)
 -
 \inf_{\substack{d\in\mathcal D\\\langle d,r_T\rangle\ge V}}\rho(d).
\]
\end{theorem}

The proof is feasible-set inclusion (Proposition~\ref{prop:library_design_proof}). It concerns the best occupancy in each hull, not the occupancy attained by greedy source switching. Hull inclusion must be established for the source libraries being compared.

\subsection{Practical Implementation: Successor Features and Alternatives}
\label{sec:successor_features}

TRAM needs repeated estimates of $Q_T^{\pi_j}(s,a)$ for many sources and actions. Successor features are useful because they turn this value computation into a fast dot product whenever rewards are linear in shared features. If $r(s,a,s')=\phi(s,a,s')^\top w$, then
\begin{align}
\psi^\pi(s,a)&=\mathbb E^\pi\!\left[
 \sum_{t\ge0}\gamma^t\phi(S_t,A_t,S_{t+1})
 \,\middle|\,S_0=s,A_0=a\right],\\
Q^\pi(s,a)&=\psi^\pi(s,a)^\top w.
\end{align}
Thus source values can be evaluated under a new target reward without retraining the source policy or running a new Bellman solve.

\begin{corollary}[Successor-feature transfer bound]
\label{cor:SF}
Under Theorem~\ref{thm:main_localized}'s assumptions and
$\|\phi(s,a,s')\|_2\le\phi_{\max}$,
\begin{equation}
\widetilde Q_T^{\pi^*}-\widetilde Q_T^{\pi_T}
 \le\frac{2\phi_{\max}}{1-\gamma}\min_j\|w_T-w_j\|_2+4Lc.
\end{equation}
\end{corollary}

The factor two accounts for the two trajectory discrepancies. Successor features are an efficient implementation route, not a definitional requirement of TRAM. When linear reward structure is unavailable, $Q_T^{\pi_j}$ can be estimated with target-task rollouts, fitted value functions, or model-based evaluation, while the source-scored deployment rule remains unchanged. The score-error bound quantifies the additional effect of evaluation error. No source policy parameters are changed after deployment in any case.

\section{Experiments}
\label{sec:experiments}
We evaluate TRAM in four settings using the zero-update source-scoring logic in Algorithm~\ref{alg:TRAM}: gridworld mechanisms, Reacher with deep successor features, Safety-Gymnasium with deployment-time spatial risk, and LLM alignment with behavioral-risk adaptation.

\paragraph{Baselines.}
We compare against risk-neutral generalized policy improvement (GPI) and risk-agnostic successor-feature transfer (SF)~\cite{sf_1,sf_2,sf_3}, risk-aware successor features (RaSF)~\cite{Gimelfarb}, and, in Safety-Gymnasium, a frozen SafeDSR-style proxy~\cite{safeDSR2024} and a one-step SFT-CoP-style dual baseline inspired by transfer-Q/source-composition decoding and constrained adaptation~\cite{Sikchi,Amrit}. These baselines test whether reward transfer alone suffices, whether variance-style risk matches deployment hazards, and whether frozen or dual-style safety proxies can match Algorithm~\ref{alg:TRAM}.

Across settings, the evaluation separates utility from the risk requested at deployment. This is important because a single return score can hide the failure mode TRAM targets: a source policy may be high-reward under the target task while accumulating the wrong kind of risk. We therefore report task performance together with the domain-specific risk measure used by the deployment score, and use the appendix diagnostics when the realized stitched behavior must be distinguished from cached source risks. The suite separates three stress tests: misspecified variance risk in small MDPs, hazards appearing after continuous-control source training, and reward-model overoptimization in decoding. In all cases, the deployed sources remain fixed.

\subsection{Gridworld: Mechanism Validation}
\label{sec:exp_gridworld}

Controlled gridworlds test whether TRAM supports expressive risk models and avoids training-time conservatism. Risk appears as spatial hazards or per-step volatility. We compare TRAM against RaSF~\cite{Gimelfarb} and risk-agnostic SF~\cite{sf_1}.

\begin{figure*}[t]
\centering
    \includegraphics[width=\textwidth]{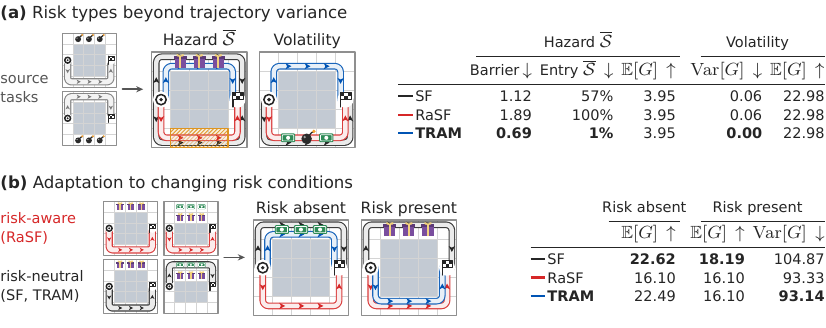}
    \caption{\textbf{Gridworld mechanism validation.} \textbf{(a)} Spatial hazards and per-step volatility expose trajectory-variance limitations; TRAM succeeds through occupancy-based risk modeling while baselines fail. \textbf{(b)} TRAM exploits high-reward paths when risk is absent and becomes conservative when risk appears, while RaSF remains uniformly conservative due to training-time risk aversion. Small grids show source policies; in (b), red sources are trained risk-aware (used by RaSF) and black sources risk-neutral (used by SF and TRAM). $G$ is the return; bold marks the best value in each column.}
    \label{fig:gridworld_combined}
\end{figure*}

Figure~\ref{fig:gridworld_combined}(a) demonstrates TRAM's expressive risk modeling. For spatial hazards, TRAM uses barrier risk to avoid danger zones while RaSF and SF fail because trajectory variance provides no spatial information. For per-step volatility, TRAM detects locally risky paths through step-level variance while trajectory-based approaches miss this fine-grained structure. Figure~\ref{fig:gridworld_combined}(b) shows adaptive risk sensitivity: TRAM exploits opportunities when risk-free but becomes conservative when risk appears, while RaSF remains uniformly conservative due to training-time risk aversion.

\subsection{Continuous Control: Reacher}
\label{sec:exp_reacher}

We evaluate scalability on Reacher (two-joint arm, 4D continuous state, MuJoCo physics). Four SFDQNs train risk-neutrally on different goals; at test time, new goals introduce danger zones. TRAM reduces barrier violations compared to risk-neutral transfer while maintaining goal performance. Successor features enable real-time adaptation through $Q_T^\pi(s,a)=\psi^\pi(s,a)^\top w_T$. This yields $\mathcal O(n|\mathcal A|d)$ dot-product scoring per step across $n$ sources and feature dimension $d$. Details are in Appendix~\ref{appendix:reacher}.

Figure~\ref{fig:continuous_results}(a) visualizes the Reacher test setting and quantitative outcomes. TRAM prioritizes avoiding the deployment-time barrier, cutting the failure rate from 85\% to 26\%, while keeping the distance-to-goal comparable to risk-neutral transfer. The comparison asks whether a fixed risk-neutral library can be re-scored when a barrier appears after training, rather than whether a new safe controller can be learned from scratch.

\subsection{Safety-Gymnasium: Deployment-Time Spatial Risk}
\label{sec:exp_safety_gym}

To test a higher-dimensional continuous-control setting, we use Safety-Gymnasium \texttt{SafetyPointGoal0-v0}~\cite{ji2023safetygymnasium}, a continuous-control safe-RL benchmark with a 16-dimensional observation representation. The target goal is fixed at $(0.7,0.7)$ and a circular deployment barrier is placed at $(0.245,0.245)$ with radius $0.18$. All methods use the same source policy library and perform no parameter updates at deployment. We evaluate 1000 episodes with horizon 1000 and $\gamma=0.99$. Additional benchmark details are in Appendix~\ref{appendix:safety_gym}.

\begin{figure*}[t]
\centering
    \includegraphics[width=0.49\textwidth]{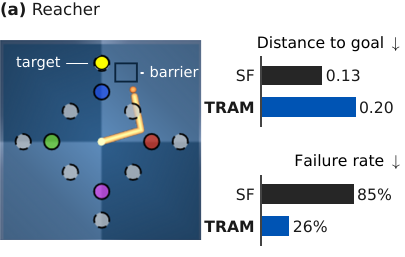}\hfill
    \includegraphics[width=0.49\textwidth]{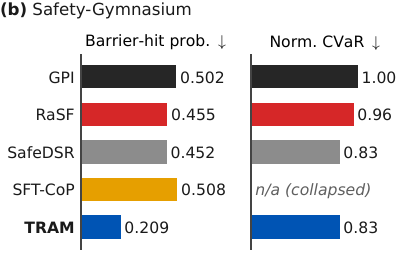}
    \caption{\textbf{Continuous-control deployment-time risk results.} \textbf{(a)} In Reacher, a new target goal and barrier are specified after source training; TRAM substantially reduces failures relative to risk-neutral SF transfer, with a modest increase in final distance to the goal. \textbf{(b)} In Safety-Gymnasium, TRAM directly scores the requested deployment barrier risk and achieves the lowest barrier-hit probability while matching the lowest normalized CVaR among valid methods.}
    \label{fig:continuous_results}
\end{figure*}

Figure~\ref{fig:continuous_results}(b) shows that TRAM reduces barrier-hit probability from 0.502 to 0.209. Variance-based RaSF improves only modestly because return variance is not aligned with the circular barrier exposure. The frozen SafeDSR proxy is close on normalized CVaR but worse on hit probability, reflecting the limitation of a non-adaptive safety proxy. SFT-CoP is omitted from the normalized-CVaR comparison because it enters a collapsed-task regime and does not provide a valid reward--risk tradeoff. We report both barrier-hit probability and normalized CVaR because they capture different failure modes: direct contact with the deployment hazard and tail exposure among risky rollouts. This directly tests the central use case: a structured spatial hazard specified only at deployment, handled by source scoring rather than retraining.

\paragraph{Constraint-aware baseline and deployment cost.}
Under a separate held-out-goal protocol (Appendix~\ref{app:sg_protocol}), constrained PPO trained on the deployment task for about 40 minutes reaches $99.7\%$ success but a barrier-hit probability of $0.24$, above its $10\%$ target. TRAM trains nothing; after re-estimating source risks from rollouts, it costs about $2.6$\,s of inference per episode and reaches $0.086$ at $c=10^{6}$, though with only $10\%$ success; smaller $c$ trades risk for success (Table~\ref{tab:sg_ppo}, Appendix~\ref{app:sg_sensitivity}).

\subsection{LLM Alignment: Behavioral Risk without Retraining}
\label{sec:exp_llm}

We also test the same principle in an inference-time LLM alignment setting~\cite{controlled_decoding,Collab}. TRAM composes two pretrained 7B source models, Zephyr-Qwen-2-7B and Dolphin-Qwen-2-7B, under a new deployment-time reward/risk specification on Berkeley Nectar~\cite{nectar}. The target reward is given by a reward model, while deployment risk is a KL-style alignment penalty to a safe reference behavior. This differs from tabular and robotic control: the action space is text generation and adaptation must occur without fine-tuning source models. Additional protocol details are in Appendix~\ref{app:LLMs}. This tests the same zero-update abstraction at the token level, as in Collab~\cite{Collab}, using external reward and risk estimators.

\begin{figure*}[t]
\centering
    \includegraphics[width=0.52\textwidth]{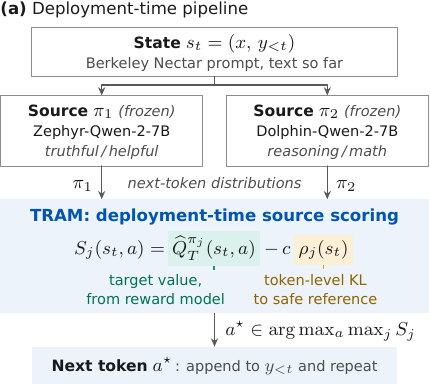}\hfill
    \includegraphics[width=0.46\textwidth]{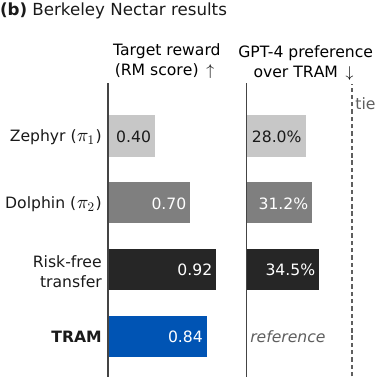}
    \caption{\textbf{LLM alignment on Berkeley Nectar.} The left panel summarizes the zero-update evaluation protocol using fixed source models, a target reward model, and a safe reference. The right panel shows that risk-free transfer maximizes the target reward-model score, while GPT-4 pairwise preferences favor TRAM overall.}
    \label{fig:llm_results_main}
\end{figure*}

Figure~\ref{fig:llm_results_main} illustrates why target-reward maximization alone can be misleading: the risk-free transfer policy obtains the highest reward-model score but is still preferred less often than TRAM by GPT-4 pairwise comparison. TRAM uses deployment-time behavioral risk to avoid overoptimizing the target reward model~\cite{gao2023overoptimization}, showing that the same source-scored rule can support risks beyond variance and low-dimensional control.

This experiment is a scope test for the source-scored principle rather than an occupancy-control claim. Unlike the gridworld and continuous-control settings, the behavioral risk is evaluated on the sources' next-token distributions, not through discounted state-action occupancies. The common structure is deployment-time scoring: fixed sources remain unchanged, the target reward model supplies the utility term, and the newly requested alignment penalty shifts which source generates the next token. Operationally, each source proposes a reusable behavior, and deployment changes only the scoring criterion. The result should therefore be read as token-level risk adaptation, not as a Markov occupancy-risk theorem for language generation. This supports the broader claim that TRAM is most useful when risk specifications change after the source behaviors are already available.

\section{Conclusion}
\label{sec:conclusion}
We introduced TRAM, a zero-update deployment-time framework that composes risk-neutral source policies using source-scored reward--risk values. The formulation treats TRAM as a tractable source-scored surrogate rather than an exact optimizer of the stitched policy's full occupancy objective, with Appendix~\ref{appendix:stitching_limitation} linking cached source risk to realized risk through a measurable stitching-mismatch term. Empirically, TRAM improves deployment safety across gridworlds, Reacher, Safety-Gymnasium, and LLM alignment without retraining. TRAM separates predeployment learning of diverse behavior from deployment-time selection of the requested risk--reward tradeoff. \textbf{Limitations.} The main limitation is source-library coverage and risk-estimation fidelity, especially if all source policies are poor. Very safe operating points can also cost task success (Appendix~\ref{app:sg_sensitivity}).

\FloatBarrier
\begin{ack}
This work was supported in part by the U.S. National Science Foundation (NSF)
under awards ECCS-1802319, 1751085, 2130706, and 2150596, and by the
U.S. Army Research Office (ARO) under Grants W911NF2310266 and W911NF261A366.
We also thank the Texas Advanced Computing Center (TACC) at The University of Texas at Austin for providing part of the computing resources used in this work.
The authors declare no competing interests.
\end{ack}

{\small
\bibliographystyle{plainnat}
\bibliography{ref}
}

\appendix
\onecolumn
\clearpage
\etocdepthtag.toc{appendix}
\etocsettagdepth{mainmatter}{none}
\etocsettagdepth{appendix}{subsection}
\etocsettocstyle{}{}
\begin{center}
{\Large\bfseries Appendix}\\[4pt]
{\normalsize TRAM: Test-Time Risk Adaptation with Mixture of Agents}
\end{center}
\label{sec:appendix}
\vspace{2pt}
\noindent\rule{\linewidth}{0.6pt}
\vspace{-4pt}
{\small\tableofcontents}
\noindent\rule{\linewidth}{0.6pt}
\vspace{6pt}

\section{Linear Programming Views of RL and Occupancy Measures}
\label{sec:occupancy}

\subsection{Policy Evaluation via Linear Programming}
\label{sec:pe_lp}

Consider a finite MDP with state space $\mathcal S$, action space $\mathcal A$, discount $\gamma\in(0,1)$, transition $p(\cdot\mid s,a)$, reward $r(s,a,s')$, and initial distribution $\mu_0$. Define
\begin{align}
\label{eq:def_rsa}
r(s,a) &:= \mathbb{E}_{s'\sim p(\cdot\mid s,a)}\!\big[r(s,a,s')\big],\\
\mathbb{P}_\pi(s'\mid s) &:= \sum_{a}\pi(a\mid s)\,p(s'\mid s,a).
\end{align}

\paragraph{Primal Q--LP (policy evaluation).}
For fixed $\pi$,
\begin{align}
\label{eq:pe_primal}
\min_{Q:\mathcal S\times\mathcal A\to\mathbb R}\quad
&(1-\gamma)\,\mathbb{E}_{s_0\sim\mu_0,\,a_0\sim\pi(\cdot\mid s_0)}\!\big[Q(s_0,a_0)\big] \nonumber\\
\text{s.t.}\quad
&Q(s,a)\ge r(s,a)+\gamma\,\mathbb E_{\substack{s'\sim p(\cdot\mid s,a)\\ a'\sim\pi(\cdot\mid s')}}\!\big[Q(s',a')\big].
\end{align}

\paragraph{Dual: discounted occupancies.}
Introducing nonnegative multipliers $d(s,a)\!\ge\!0$ yields
\begin{align}
\max_{d\ge 0}\quad &\sum_{s,a} d(s,a)\,r(s,a) \\
\text{s.t.}\quad
&d(s,a)=(1-\gamma)\mu_0(s)\pi(a\mid s) \nonumber\\
&\qquad\quad +\gamma\!\sum_{s',a'}\!d(s',a')p(s\mid s',a')\pi(a\mid s),\quad \forall (s,a). \nonumber
\end{align}
At optimality $d=d^\pi$. The normalized discounted occupancy is
\begin{keybox}
\begin{align}
\label{eq:disc_occupancy}
d^\pi(s,a)=(1-\gamma)\sum_{t=0}^{\infty}\gamma^t\,\Pr_\pi(S_t=s,A_t=a\mid S_0\sim\mu_0).
\end{align}
\end{keybox}
Strong duality gives $(1-\gamma)\,\mathbb E_{\mu_0,\pi}[Q^\pi(S_0,A_0)]=\langle d^\pi,r\rangle$. If $\nu(s):=\sum_a d(s,a)>0$, a policy is recovered by $\pi(a\mid s)=d(s,a)/\nu(s)$.

\subsection{Risk-Regularized Control via Occupancy Optimization}
\label{sec:control_lp}

Define the \emph{occupancy polytope}
\begin{keybox}
\begin{align}
\label{eq:occupancy_polytope}
\mathcal{D}
=\Big\{
d\!\in\!\mathbb{R}_{\ge 0}^{\mathcal S\times\mathcal A}:\;
\underbrace{\sum_a d(s,a)}_{\nu(s)}
&=(1-\gamma)\mu_0(s) \nonumber\\
&\quad+\gamma\!\sum_{s',a'}\! d(s',a')\,p(s\mid s',a'),\ \forall s
\Big\}.
\end{align}
\end{keybox}
Given a risk functional $\rho:\mathcal D\!\to\!\mathbb R\cup\{+\infty\}$ with nonempty finite domain, the penalized control problem for $c_{\mathrm{occ}}>0$ is
\begin{align}
\label{eq:v_lp_risk}
\max_{d\in\mathcal D}\ \langle d,r\rangle - c_{\mathrm{occ}}\,\rho(d),\qquad c_{\mathrm{occ}}>0,
\end{align}
whose optimizer, when it exists, induces $\pi^\star(a\mid s)=d^\star(s,a)/\nu^\star(s)$ wherever $\nu^\star(s)>0$; actions at zero-occupancy states can be chosen arbitrarily. At $c_{\mathrm{occ}}=0$, use the ordinary reward-only occupancy problem without evaluating $0\cdot\rho(d)$ at infinite risks. To match the unnormalized-return objective in Eq.~\eqref{eq:penalized_surrogate}, use $c_{\mathrm{occ}}=(1-\gamma)c$.
For one-step functionals, we use the occupancy-weighted expectation
\begin{align}
\label{eq:Ed_def}
\mathbb{E}_d[\phi(s,a,s')]
:=\sum_{s,a}d(s,a)\sum_{s'}p(s'\mid s,a)\,\phi(s,a,s').
\end{align}

\subsection{Risk Functionals on Occupancies: Compact Catalog}
\label{sec:risk_functionals}

All functionals below depend \emph{only} on $d$ (and possibly a fixed reference $\bar d$) and thus fit \eqref{eq:v_lp_risk}.

\paragraph{(R1) Set-occupancy constraints (barrier / chance proxy).}
For danger set $\overline{\mathcal S}$ and tolerance $\delta\!\in\!(0,1)$,
\begin{align}
\label{eq:barrier_risk}
\rho_{\text{set}}(d)\;=\;-\log\!\left(\delta-\sum_{s\in\overline{\mathcal S}}\nu(s)\right),
\quad \text{valid when }\sum_{s\in\overline{\mathcal S}}\nu(s)<\delta,
\end{align}
which penalizes discounted state-mass in $\overline{\mathcal S}$. This is an exposure surrogate, not the probability of ever hitting the danger set.

\paragraph{(R2) Per-step reward variance (local volatility).}
Let $\tilde r(s,a):=\mathbb E_{s'\mid s,a}[r(s,a,s')^2]$. Since $\mathbb E_d[r]=\langle d,r\rangle$ under the normalized occupancy,
\begin{align}
\label{eq:variance_rho}
\rho_{\text{var}}(d)\;=\;\langle d,\tilde r\rangle-\big(\langle d,r\rangle\big)^2,
\end{align}
the variance of a single reward drawn from the occupancy-weighted transition distribution. It can be positive for deterministic trajectories whose rewards vary across states. It is generally concave, not convex, as a function of $d$.

\paragraph{(R3) $f$-divergences to a safe reference (behavioral proximity).}
For a reference occupancy $\bar d$ with matching support, the KL case is
\begin{align}
\label{eq:kl_risk}
\rho_{\text{KL}}(d)=\sum_{s,a} d(s,a)\log\!\frac{d(s,a)}{\bar d(s,a)}.
\end{align}
Other $f$-divergences, such as $\chi^2$, also depend only on the two occupancies. R\'enyi divergences provide another family of occupancy comparisons, with convexity depending on the order and argument.

\paragraph{(R4) Integral probability metrics (IPMs), e.g., MMD.}
Let $\Phi$ be a feature map into an RKHS $\mathcal H$:
\begin{align}
\rho_{\text{MMD}}(d)=\Big\|\sum_{s,a}d(s,a)\Phi(s,a)-\sum_{s,a}\bar d(s,a)\Phi(s,a)\Big\|_{\mathcal H}.
\end{align}
This provides a nonparametric proximity to a safe/reference occupancy.

\paragraph{(R5) Entropic (exponential) per-step risk.}
Given a step risk cost $g(s,a,s')$ and $\theta>0$,
\begin{align}
\label{eq:entropic}
\rho_{\text{ent}}(d)\;=\;\tfrac{1}{\theta}\log\!\Big(\mathbb E_d\big[e^{\theta\,g(s,a,s')}\big]\Big),
\end{align}
a tail-sensitive single-step penalty. For fixed $g$, the positive expectation is affine in $d$, so its logarithm is concave in $d$. Convexity in the losses must not be confused with convexity in occupancy.

\paragraph{(R6) Energy / actuation budgets (control cost).}
With an action embedding $u(s,a)\in\mathbb R^m$ and $G\succeq0$,
\begin{align}
\rho_{\text{energy}}(d)=\sum_{s,a} d(s,a)\,u(s,a)^\top G\,u(s,a),
\end{align}
which constrains power, wear, or acceleration magnitudes.

\paragraph{(R7) Transition smoothness / jerk (temporal regularity).}
Let $\varphi(s,a,s')$ encode finite differences (e.g., $\Delta s$); a control increment or jerk requires the relevant control history to be included in the state. Define
\begin{align}
\rho_{\text{smooth}}(d)=\mathbb E_d\big[\|\varphi(s,a,s')\|_2^2\big],
\end{align}
penalizing oscillatory behaviors via occupancy-weighted transitions.

\paragraph{(R8) Reach--avoid proxy (linear temporal-logic surrogate).}
For target set $\mathcal T$ and avoid set $\overline{\mathcal S}$,
\begin{align}
\rho_{\text{RA}}(d)=\alpha\,\sum_{s\in\overline{\mathcal S}}\nu(s)
+\beta\,\Big(1-\sum_{s\in\mathcal T}\nu(s)\Big),\qquad \alpha,\beta\ge 0,
\end{align}
a linear-in-$d$ proxy for STL/LTL reach--avoid objectives.

\paragraph{Design notes.}
(i) Each $\rho$ above is a function of $(d,\bar d)$ only; no return distributions or value iterations are needed.
(ii) KL, the IPM norm, and the log barrier on its domain are convex in $d$; fixed-cost energy, smoothness, and reach--avoid penalties are linear in $d$. These choices make the penalized occupancy objective concave. The variance and entropic examples do not share that property.
(iii) The analysis requires a finite Lipschitz constant on the occupancies being compared. Bounded linear costs satisfy this condition. A log barrier requires slack $\delta-\sum_{s\in\overline{\mathcal S}}\nu(s)\ge\varepsilon>0$, giving $L\le1/\varepsilon$. KL has a bounded gradient on domains with density ratios bounded above and away from zero; positive reference mass alone does not make KL globally Lipschitz up to the boundary. A source satisfying these domain restrictions does not imply that the stitched policy satisfies them.

\subsection{Connection to TRAM}
\label{sec:connection_tram}

In TRAM, each risk-neutral source $\pi_j$ is evaluated on the \emph{target} to obtain its occupancy $d^{\pi_j}$ and risk score $\rho(d^{\pi_j})$. The risk-adjusted utility
\begin{align}
\label{eq:tram_signal}
\mathcal U_j(s,a)=Q^{\pi_j}(s,a)-c\,\rho\!\big(d^{\pi_j}\big)
\end{align}
drives per-state action selection (Eq.~\ref{eq:TRAM_policy}) without any parameter updates. Risk is specified through $\rho$ and can be changed at deployment. Diverse sources can improve coverage, but risk-neutral training does not automatically maximize coverage of $\mathcal D$; the source-hull comparison in Theorem~\ref{thm:library_design} requires explicit hull inclusion.

\section{Proofs and Technical Guarantees}
\label{app:proofs}

\begin{assumptionbox}
\textbf{Analysis setting.}
We consider finite discounted MDPs with bounded expected one-step rewards, common transition kernel $p$, discount $\gamma\in(0,1)$, and a fixed initial distribution $\mu_0$. Policies are stationary. A superscript $*$ denotes a policy optimal from every state, not merely for one initial distribution. Each source policy $\pi_j$ is statewise optimal for its source reward $r_j$ under $p$, so $Q_j^{\pi_j}=Q_j^*$. Unless estimation error is stated explicitly, action values and source risks are exact. Any hard source screen leaves a nonempty, fixed library, and all source indices below refer to that retained library.
\end{assumptionbox}

The normalized discounted occupancy is
\[
d^\pi(s,a)=(1-\gamma)\sum_{t\ge0}\gamma^t
 \Pr_\pi(S_t=s,A_t=a\mid S_0\sim\mu_0).
\]
Thus $\|d^\pi\|_1=1$ and $\langle d^\pi,r\rangle=(1-\gamma)J_r(\pi)$, where $J_r$ is the unnormalized discounted return used in the main text. Conditional occupancies $d^{\pi\mid s,a}$ fix the initial state--action pair and satisfy
\[
Q_r^\pi(s,a)=\frac{\langle d^{\pi\mid s,a},r\rangle}{1-\gamma}.
\]
Source risks use the global occupancy from $\mu_0$, not this conditional occupancy. Write
\[
\rho_j=\rho(d^{\pi_j}),\qquad
\widetilde Q_i^\pi(s,a)=Q_i^\pi(s,a)-c\rho(d^\pi),\qquad c\ge0.
\]
The risk-adjusted quantity is a score, not a Bellman value function. Whenever a Lipschitz bound is invoked, $\rho$ is finite and $L$-Lipschitz in $\ell_1$ on a domain containing all occupancies being compared. Convexity is needed only for the source-hull bridge in Appendix~\ref{app:stitching_diagnostics}.

For tasks $i,j$, define
\[
\Delta_\pi^{(i,j)}(s,a)=
\mathbb E_\pi\!\left[
 \sum_{t\ge0}\gamma^t|r_i(S_t,A_t)-r_j(S_t,A_t)|
 \,\middle|\,S_0=s,A_0=a\right].
\]

\subsection{Basic Value-Difference Lemmas}

\begin{lemma}[Fixed-policy value difference]\label{lem:fixed-local}
For any tasks $i,j$, policy $\pi$, and $(s,a)$,
\[
|Q_i^\pi(s,a)-Q_j^\pi(s,a)|\le\Delta_\pi^{(i,j)}(s,a).
\]
\end{lemma}
\begin{proof}
Under common dynamics, the two values integrate different rewards along the same trajectory distribution. Subtract their discounted-return expansions, apply the triangle inequality inside the expectation, and sum. Bounded rewards justify the interchange of expectation and the discounted sum.
\end{proof}

\begin{lemma}[Optimal--optimal comparison]\label{lem:opt-opt}
Let $\pi_i^*,\pi_j^*$ be statewise optimal for tasks $i,j$. Then
\begin{align*}
-\Delta_{\pi_j^*}^{(i,j)}(s,a)
&\le Q_i^*(s,a)-Q_j^*(s,a)
\le \Delta_{\pi_i^*}^{(i,j)}(s,a),\\
|Q_i^*(s,a)-Q_j^*(s,a)|
&\le\max\{\Delta_{\pi_i^*}^{(i,j)}(s,a),
               \Delta_{\pi_j^*}^{(i,j)}(s,a)\}.
\end{align*}
\end{lemma}
\begin{proof}
Statewise optimality gives $Q_j^*\ge Q_j^{\pi_i^*}$, so
\[
Q_i^*-Q_j^*
 \le Q_i^{\pi_i^*}-Q_j^{\pi_i^*}
 \le\Delta_{\pi_i^*}^{(i,j)}.
\]
Exchanging $i$ and $j$ gives $Q_j^*-Q_i^*\le\Delta_{\pi_j^*}^{(i,j)}$. These are bounds on opposite signs of the difference; the absolute-value bound uses their maximum.
\end{proof}

\begin{lemma}[Transfer loss of a source policy]\label{lem:opt-eval}
For any source policy $\pi_j$ and target task $i$,
\[
0\le Q_i^*(s,a)-Q_i^{\pi_j}(s,a)
 \le C_j^{(i)}(s,a),
\quad
C_j^{(i)}=\Delta_{\pi_i^*}^{(i,j)}+\Delta_{\pi_j}^{(i,j)}.
\]
\end{lemma}
\begin{proof}
Insert the source-task values:
\begin{align*}
Q_i^*-Q_i^{\pi_j}
={}&(Q_i^{\pi_i^*}-Q_j^{\pi_i^*})
 +(Q_j^{\pi_i^*}-Q_j^{\pi_j})
 +(Q_j^{\pi_j}-Q_i^{\pi_j}).
\end{align*}
The first and third terms are at most the respective discrepancies by Lemma~\ref{lem:fixed-local}. The middle term is nonpositive because $\pi_j$ is optimal for task $j$: $Q_j^{\pi_j}=Q_j^*\ge Q_j^{\pi_i^*}$. The lower bound follows from target-task optimality.
\end{proof}

\begin{lemma}[Occupancy diameter]\label{lem:tv}
For any policies $\pi,\pi'$ evaluated with the same occupancy convention,
$\|d^\pi-d^{\pi'}\|_1\le2$.
\end{lemma}
\begin{proof}
Both occupancies are probability distributions, and the triangle inequality bounds their distance by the sum of their $\ell_1$ norms.
\end{proof}

\begin{lemma}[Lipschitz risk gap]\label{lem:rho-lip}
For occupancies in the Lipschitz domain,
$|\rho(d^\pi)-\rho(d^{\pi'})|\le L\|d^\pi-d^{\pi'}\|_1\le2L$.
\end{lemma}
\begin{proof}
Apply the Lipschitz assumption and Lemma~\ref{lem:tv}.
\end{proof}

\subsection{Generalized Improvement with Source-Level Penalties}

The following argument extends the Bellman monotonicity reasoning behind generalized policy improvement~\cite{sf_1}; its constant shift is essential when source risks can be negative.

\begin{definition}[Bellman operators]\label{def:bellman}
For the target reward $r_T$, define
\[
(T_T^\pi Q)(s,a)=r_T(s,a)+\gamma
 \mathbb E_{s'\sim p(\cdot\mid s,a),\,b\sim\pi(\cdot\mid s')}Q(s',b).
\]
It is monotone and a $\gamma$-contraction, with fixed point $Q_T^\pi$.
The optimality operator replaces the inner policy expectation by $\max_b$.
\end{definition}

Let $\rho_{\min}=\min_j\rho_j$ and
\[
\alpha_j=c(\rho_j-\rho_{\min})\ge0,\qquad
F(s,a)=\max_j\{Q_T^{\pi_j}(s,a)-\alpha_j\}.
\]
Adding the common constant $c\rho_{\min}$ to all original TRAM scores does not change action selection, so TRAM is greedy with respect to $F$.

\begin{lemma}[Shifted source-score improvement]\label{lem:lift}
For exact source values and finite source risks, the greedy TRAM policy satisfies
\[
Q_T^{\pi_T}(s,a)\ge F(s,a)
 =\max_j\{Q_T^{\pi_j}(s,a)-c(\rho_j-\rho_{\min})\}.
\]
No sign or convexity assumption on $\rho$ is required.
\end{lemma}
\begin{proof}
For every source $j$, monotonicity and its Bellman equation imply
\begin{align*}
T_T^{\pi_T}F=T_T^*F
&\ge T_T^{\pi_j}F\\
&\ge T_T^{\pi_j}(Q_T^{\pi_j}-\alpha_j)\\
&=Q_T^{\pi_j}-\gamma\alpha_j
 \ge Q_T^{\pi_j}-\alpha_j .
\end{align*}
Taking the maximum over $j$ gives $T_T^{\pi_T}F\ge F$.
Iterating this monotone contraction and taking the limit proves the claim.
The identity $T_T^{\pi_j}(Q-\alpha)=T_T^{\pi_j}Q-\gamma\alpha$ is why a nonnegative shifted penalty is used.
\end{proof}

\subsection{Localized Transfer Bounds}

\begin{lemma}[Risk-adjusted comparison with a source]\label{lem:tildeQ-RN}
For any source $j$,
\[
|\widetilde Q_T^{\pi^*}-\widetilde Q_T^{\pi_j}|
 \le C_j^{(T)}+2Lc,
\]
where $\pi^*$ is target-optimal and all quantities are evaluated at the same $(s,a)$.
\end{lemma}
\begin{proof}
By Lemma~\ref{lem:opt-eval}, the nonnegative reward-value gap is at most $C_j^{(T)}$.
The absolute difference of the two risk penalties is at most $2Lc$ by Lemma~\ref{lem:rho-lip}. Apply the triangle inequality.
\end{proof}

\begin{theorem}[Localized bounds for the deployed policy]\label{thm:main-RN}
Under the analysis assumptions, set $C_j=C_j^{(T)}$. Then
\begin{align}
Q_T^*-Q_T^{\pi_T}
&\le \min_j\{C_j+c(\rho_j-\rho_{\min})\}
 \le\min_j C_j+2Lc,\label{eq:reward_transfer_proof}\\
\widetilde Q_T^{\pi^*}-\widetilde Q_T^{\pi_T}
&\le\min_j C_j+4Lc.\label{eq:risk_transfer_proof}
\end{align}
\end{theorem}
\begin{proof}
Lemma~\ref{lem:lift} gives, for every $j$,
\[
Q_T^*-Q_T^{\pi_T}
 \le Q_T^*-Q_T^{\pi_j}+c(\rho_j-\rho_{\min})
 \le C_j+c(\rho_j-\rho_{\min}).
\]
Take the minimum and use $\rho_j-\rho_{\min}\le2L$.
For the second inequality, use the actual deployed-policy risk:
\[
\widetilde Q_T^{\pi^*}-\widetilde Q_T^{\pi_T}
 =Q_T^*-Q_T^{\pi_T}
   +c\{\rho(d^{\pi_T})-\rho(d^{\pi^*})\}.
\]
The last term is at most $2Lc$. This step does not identify the deployed risk with any cached source risk.
\end{proof}

\paragraph{Scope of the comparison.}
The comparator is a risk-neutral target optimum. These bounds do not establish optimality for the risk-regularized control problem or satisfaction of a hard risk threshold. They apply to the deployed policy only when its actual occupancy belongs to the stated Lipschitz domain. The convex source-hull bridge is a separate diagnostic, not a substitute for this domain condition.

\subsection{Main-Text Corollaries and Source-Library Design}
\label{app:main_text_proofs}

\begin{proposition}[Main-text localized guarantee]\label{prop:localized_main_proof}
Theorem~\ref{thm:main_localized} follows with
$C_j=\Delta_{\pi^*}^{(T,j)}+\Delta_{\pi_j}^{(T,j)}$.
\end{proposition}
\begin{proof}
Apply Eq.~\eqref{eq:risk_transfer_proof}. Both trajectory discrepancies are required to compare a target-optimal policy with a source policy.
\end{proof}

\begin{proposition}[Reward-only corollary]\label{prop:standard_corollary_proof}
Corollary~\ref{cor:standard_performance} has penalty term $2Lc$.
\end{proposition}
\begin{proof}
Use Eq.~\eqref{eq:reward_transfer_proof} directly. Removing the risk-adjusted terms from a looser bound would give an unnecessary additional penalty.
\end{proof}

\begin{proposition}[Successor-feature corollary]\label{prop:sf_corollary_proof}
If $r_k(s,a,s')=\phi(s,a,s')^\top w_k$ with
$\|\phi(s,a,s')\|_2\le\phi_{\max}$, then
\[
C_j(s,a)\le
 \frac{2\phi_{\max}}{1-\gamma}\|w_T-w_j\|_2.
\]
\end{proposition}
\begin{proof}
The difference of the expected one-step rewards is bounded by
$\phi_{\max}\|w_T-w_j\|_2$ using Cauchy--Schwarz and the common transition kernel. Each of the two trajectory discrepancies is therefore at most this quantity times $(1-\gamma)^{-1}$.
\end{proof}

\begin{proposition}[Conditional source-hull dominance]\label{prop:library_design_proof}
Let $S_\lambda\subseteq\mathcal D$ be occupancy libraries and assume
$\operatorname{co}(S_\lambda)\subseteq\operatorname{co}(S_0)$ for every $\lambda\ge0$.
Let every hull intersect $\{d:\langle d,r_T\rangle\ge V\}$.
For any nonempty class $\mathcal R_L$ of finite $L$-Lipschitz risks on $\mathcal D$, define the nonnegative regret
\[
\mathcal E_{\rho,V}(S)=
 \inf_{\substack{d\in\operatorname{co}(S)\\\langle d,r_T\rangle\ge V}}\rho(d)
 -
 \inf_{\substack{d\in\mathcal D\\\langle d,r_T\rangle\ge V}}\rho(d).
\]
Then
\[
\sup_{\rho\in\mathcal R_L}\mathcal E_{\rho,V}(S_0)
 =\inf_{\lambda\ge0}\sup_{\rho\in\mathcal R_L}\mathcal E_{\rho,V}(S_\lambda).
\]
\end{proposition}
\begin{proof}
Hull inclusion enlarges the feasible set of the first infimum for $S_0$ and can only decrease it. The second infimum is common to all libraries, so
$\mathcal E_{\rho,V}(S_0)\le\mathcal E_{\rho,V}(S_\lambda)$.
Taking suprema preserves this inequality; inclusion of $\lambda=0$ gives equality after the infimum over $\lambda$. Feasibility and the Lipschitz diameter bound make the regrets finite.
\end{proof}

This is a conditional set-inclusion result. Risk-neutral training alone does not imply the required inclusion, and the result concerns best achievable risk in a source hull, not the performance of the greedy TRAM rule.

\subsection{Approximate Source Scores}
\label{app:score_error}

\begin{proposition}[Uniform score-error bound]\label{prop:score_error}
Suppose $\widehat\pi$ is greedy with respect to
$\widehat M(s,a)=\max_j\{\widehat Q_j(s,a)-c\widehat\rho_j\}$ and
\[
\sup_{j,s,a}
|(\widehat Q_j(s,a)-c\widehat\rho_j)
 -(Q_T^{\pi_j}(s,a)-c\rho_j)|\le\eta .
\]
For a fixed nonempty source library,
\[
Q_T^{\widehat\pi}\ge F-\frac{2\gamma\eta}{1-\gamma}.
\]
Consequently, each bound in Theorem~\ref{thm:main-RN} holds for $\widehat\pi$ after adding $2\gamma\eta/(1-\gamma)$ to its right-hand side, provided the risk-domain assumption includes $d^{\widehat\pi}$.
\end{proposition}
\begin{proof}
Taking a maximum cannot increase the uniform error. Since $F$ differs from the true max-score by a constant, greedy selection under $\widehat M$ satisfies
$F(s,\widehat\pi(s))\ge\max_a F(s,a)-2\eta$.
Thus
\[
T_T^{\widehat\pi}F\ge T_T^*F-2\gamma\eta\ge F-2\gamma\eta.
\]
Set $h=2\gamma\eta/(1-\gamma)$.
Constant-shift equivariance gives
$T_T^{\widehat\pi}(F-h)\ge F-2\gamma\eta-\gamma h=F-h$.
Monotone contraction yields $Q_T^{\widehat\pi}\ge F-h$.
The transfer bounds then follow exactly as in Theorem~\ref{thm:main-RN}.
\end{proof}
The bound is for action values with the initial action fixed. If
$\|\widehat Q_j-Q_T^{\pi_j}\|_\infty\le\epsilon_Q$ and
$|\widehat\rho_j-\rho_j|\le\epsilon_\rho$, one may take
$\eta=\epsilon_Q+c\epsilon_\rho$. Screening errors require a separate guarantee that the retained library is appropriate; this proposition only analyzes scoring within a fixed retained library.

\section{Stitching Mismatch and Approximate-Dynamics Extensions}
\label{appendix:stitching_limitation}

\subsection{Stitching Mismatch and the Source-Hull Bridge}
\label{app:stitching_diagnostics}

The action rule in Eq.~\eqref{eq:TRAM_policy} can select different sources across states, so $d^{\pi_T}$ need not equal any source occupancy or their selected mixture. Fix any weights $\omega_j\ge0$ with $\sum_j\omega_j=1$ and define
\[
\bar d_T=\sum_j\omega_jd^{\pi_j},
\qquad
\epsilon_{\mathrm{stitch}}=\|d^{\pi_T}-\bar d_T\|_1.
\]
One possible choice is the normalized discounted frequency of each selected source along deployed trajectories, with a fixed tie-breaking rule. The proposition also allows other mixture weights; it does not identify source-selection frequencies with mixture occupancy weights.

\begin{proposition}[Source-hull realized-risk bridge]
\label{prop:stitch_bound}
If $\rho$ is convex and $L$-Lipschitz on a convex domain containing the source occupancies and $d^{\pi_T}$, then
\begin{equation}
\label{eq:stitch_bound}
\rho(d^{\pi_T})\le
 \sum_j\omega_j\rho(d^{\pi_j})+L\epsilon_{\mathrm{stitch}}.
\end{equation}
\end{proposition}
\begin{proof}
Lipschitzness gives $\rho(d^{\pi_T})\le\rho(\bar d_T)+
L\|d^{\pi_T}-\bar d_T\|_1$. Jensen's inequality gives
$\rho(\bar d_T)\le\sum_j\omega_j\rho(d^{\pi_j})$. Combine the two inequalities.
\end{proof}

For fixed weights, the mismatch term is zero exactly when $d^{\pi_T}=\bar d_T$. Membership in the source hull alone is insufficient to make a particular frequency-weighted mismatch zero. Minimizing the distance over all mixture weights gives the distance to the hull; a large distance for one chosen mixture does not prove a large distance to the entire hull.

The bridge does not apply to every occupancy functional. In particular, the variance and entropic examples in Appendix~\ref{sec:risk_functionals} are generally concave in occupancy. Without convexity, Lipschitzness still gives
$\rho(d^{\pi_T})\le\rho(\bar d_T)+L\epsilon_{\mathrm{stitch}}$, but the source-risk average cannot replace $\rho(\bar d_T)$.

\paragraph{Estimation and certification.}
Finite-rollout mismatch estimates are diagnostics unless their errors are bounded. For any chosen weights, if occupancy estimates satisfy
$\|\widehat d_T-d^{\pi_T}\|_1\le\zeta_T$ and
$\|\widehat d_j-d^{\pi_j}\|_1\le\zeta_j$, the triangle inequality gives
\[
\epsilon_{\mathrm{stitch}}\le
 \left\|\widehat d_T-\sum_j\omega_j\widehat d_j\right\|_1
 +\zeta_T+\sum_j\omega_j\zeta_j.
\]
If source-risk estimates also satisfy
$|\widehat\rho_j-\rho(d^{\pi_j})|\le\xi_j$, a valid upper bound is
\[
\rho(d^{\pi_T})\le
 \sum_j\omega_j(\widehat\rho_j+\xi_j)
 +L\left(
 \left\|\widehat d_T-\sum_j\omega_j\widehat d_j\right\|_1
 +\zeta_T+\sum_j\omega_j\zeta_j\right).
\]
When the error bounds hold jointly with high probability, so does this inequality. Continuous-state applications need an appropriate occupancy representation and associated approximation bounds; empirical rollouts alone do not provide an exact certificate.

\subsection{Approximate Dynamics and Source Evaluation}
\label{app:dynamics_mismatch}

The common-dynamics analysis evaluates every source in the target MDP, where each source is assumed optimal for its own reward $r_j$. If a different model is also used to evaluate the target reward, its error affects greedy action selection as follows.

\begin{theorem}[Model-error extension]
\label{thm:dynamics_mismatch}
Suppose source $j$ is evaluated exactly in model $(\widehat p_j,\widehat r_{T,j})$, yielding $\widehat Q_j=Q_{\widehat p_j,\widehat r_{T,j}}^{\pi_j}$, while the target uses $(p_T,r_T)$. Assume uniformly over $j,s,a$ that
\[
\|\widehat p_j(\cdot\mid s,a)-p_T(\cdot\mid s,a)\|_{\mathrm{TV}}\le\delta_p,
\quad
|\widehat r_{T,j}(s,a)-r_T(s,a)|\le\delta_r,
\quad
\|\widehat Q_j\|_\infty\le B.
\]
If source-risk errors are at most $\epsilon_\rho$, define
\[
\eta_{\mathrm{model}}=
 \frac{\delta_r+2\gamma B\delta_p}{1-\gamma}+c\epsilon_\rho.
\]
Under the remaining risk-domain assumptions of Theorem~\ref{thm:main_localized}, the policy $\widehat\pi$ greedy in these model-based scores obeys
\begin{align}
\label{eq:dynamics_mismatch_bound}
Q_T^*-Q_T^{\widehat\pi}
&\le\min_j C_j+2Lc+
 \frac{2\gamma}{1-\gamma}\eta_{\mathrm{model}},\\
\widetilde Q_T^{\pi^*}-\widetilde Q_T^{\widehat\pi}
&\le\min_j C_j+4Lc+
 \frac{2\gamma}{1-\gamma}\eta_{\mathrm{model}}.
\end{align}
All $C_j$, source occupancies, and true risks in these comparisons use the target dynamics.
\end{theorem}
\begin{proof}
Let $\widehat V_j(s)=\sum_a\pi_j(a\mid s)\widehat Q_j(s,a)$, so
$\|\widehat V_j\|_\infty\le B$. Total variation yields
\[
|\mathbb E_{p_T}\widehat V_j-\mathbb E_{\widehat p_j}\widehat V_j|
 \le2B\delta_p.
\]
Since $\widehat Q_j$ is the fixed point of the model Bellman operator, its residual under the target operator for $\pi_j$ is at most $\delta_r+2\gamma B\delta_p$. Contraction therefore gives
\[
\|Q_T^{\pi_j}-\widehat Q_j\|_\infty
 \le\frac{\delta_r+2\gamma B\delta_p}{1-\gamma}.
\]
Adding the source-risk error gives a uniform score error of at most $\eta_{\mathrm{model}}$. Proposition~\ref{prop:score_error} then proves both bounds.
\end{proof}

If rewards and risks are evaluated exactly and only the transition model differs, the added action-value term is
$4\gamma^2B\delta_p/(1-\gamma)^2$.
Both the value-evaluation error and the effect of choosing actions with erroneous scores must be accounted for.
Additional critic approximation error can be added to $\eta_{\mathrm{model}}$.
For transition-dependent rewards, $\delta_r$ accounts for the change in their expected one-step value; it is zero only when these expectations are evaluated correctly.

\section{Continuous-Control and Safety-Gymnasium Experimental Details}
\label{app:continuous_details}

This appendix records implementation details for the continuous-control results in Fig.~\ref{fig:continuous_results} (Appendices~\ref{appendix:reacher}--\ref{appendix:safety_gym}) and reports additional Safety-Gymnasium experiments: a constraint-aware baseline and deployment costs, sensitivity to $c$, an alternative risk functional, and the cost of occupancy estimation (Appendices~\ref{app:sg_protocol}--\ref{app:sg_occupancy}).

\subsection{Reacher Implementation Details}
\label{appendix:reacher}

\textbf{Environment.}
\textsc{Reacher} (MuJoCo~\cite{MuJoCo}) uses a planar 2-DOF arm with continuous joint-angle and velocity states. Following successor-feature transfer protocols~\cite{sf_1,Gimelfarb,Amrit_sfdqn}, torques are discretized to nine actions corresponding to minimum, zero, or maximum torque on each joint. Rewards combine negative distance-to-goal with a small control penalty.

\textbf{Source library.}
Four risk-neutral Successor-Feature DQNs (SFDQNs) are trained on different source goals. Each source learns successor features $\psi^{\pi_j}$ so that the target value can be evaluated by the dot product
\[
Q_T^{\pi_j}(s,a)=\psi^{\pi_j}(s,a)^\top w_T.
\]
At deployment, the source parameters are frozen. TRAM only evaluates target values and the specified risk term for the fixed source library.

\textbf{Deployment-time risk.}
The target task changes the goal and introduces a barrier region after source training. We use a barrier-exposure risk computed from target rollouts or cached occupancy estimates, with tradeoff weight $c=5$. Thus the deployment-time computation is source scoring, not retraining or re-solving the target MDP. A failure is recorded when the arm's tip enters the barrier region, and the test is repeated 100 times.

\textbf{Runtime.}
For $n$ sources, $|\mathcal A|$ discrete actions, and feature dimension $d$, action selection costs $\mathcal O(n|\mathcal A|d)$ per state when successor-feature dot products are used. The risk term is a cached source-level lookup, so the online overhead over risk-neutral SF transfer is small.

\subsection{Safety-Gymnasium Benchmark Details}
\label{appendix:safety_gym}

We use Safety-Gymnasium \texttt{SafetyPointGoal0-v0}~\cite{ji2023safetygymnasium}. The target goal is fixed at $(0.7,0.7)$ and the deployment risk is a circular barrier centered at $(0.245,0.245)$ with radius $0.18$. The observation representation is 16-dimensional. We evaluate 1000 episodes with horizon 1000 and discount $\gamma=0.99$.

All methods use the same fixed source library and perform no deployment-time parameter updates. Risk-neutral GPI scores target reward only. RaSF uses a variance-style surrogate. The SafeDSR-style proxy is frozen and therefore cannot fully rescore behavior under the newly requested barrier risk. The SFT-CoP-style baseline enters a collapsed-task regime in this benchmark, so it is shown for barrier-hit probability in the main figure but omitted from the normalized-CVaR comparison.

The normalized-CVaR exposure in Fig.~\ref{fig:continuous_results}(b) rescales exposure relative to the risk-neutral GPI baseline so that lower values indicate safer behavior under the requested deployment-time barrier risk. This normalization is used only for visual comparability across methods.

\subsection{Additional Safety-Gymnasium Experiments: Protocol}
\label{app:sg_protocol}

The experiments in Appendices~\ref{app:sg_ppo}--\ref{app:sg_occupancy} use a separate protocol from Fig.~\ref{fig:continuous_results}(b), with a different held-out goal, so their numbers should not be compared with that figure directly.

\textbf{Task.} \texttt{SafetyPointGoal0-v0} with held-out target goal $(0.5,0.5)$ and goal tolerance $0.02$. The deployment barrier is a circle centered at $(0.245,0.245)$ with radius $0.18$ unless stated otherwise. Each evaluation uses 300 episodes with horizon $1{,}000$ and $\gamma=0.99$. Success means reaching the goal; barrier-hit probability is the fraction of episodes that enter the barrier. All methods share one evaluator (goal, barrier geometry, horizon, success and hit detectors).

\textbf{Source library and risk.} The library contains 39 frozen SF-DQN source policies trained risk-neutrally on goals in the cardinal and diagonal directions; none is trained on the target goal. For each source we cache a policy-level risk, its barrier-hit probability under the deployment barrier, estimated offline from rollouts of the frozen policy. TRAM uses the dynamic per-state rule of Algorithm~\ref{alg:TRAM}: at every state it scores each source by its successor-feature value minus $c$ times the cached risk and executes the best action.

\textbf{Scale of $c$.} The SF-DQN action values in this environment range from about $10^2$ to $10^5$, while the cached risk is a probability. Penalty weights that change the selected source are therefore of order $10^4$ to $10^6$; $c$ is meaningful only relative to the value scale of the library.

\textbf{Hardware.} These experiments were run on NVIDIA GH200 Grace Hopper nodes, one node per job, on the node's Grace CPU (Appendix~\ref{app:repro}). Wall-clock times exclude scheduler queueing.

\subsection{Constraint-Aware Baseline and Deployment Cost}
\label{app:sg_ppo}

\textbf{Constrained PPO.} As a constraint-aware reference, we train a PPO policy directly on the deployment specification (goal $(0.5,0.5)$, barrier radius $0.18$). It uses a separate risk critic and an adaptive Lagrange multiplier that targets a barrier-hit probability of $0.10$; the target is annealed from $0.30$ to $0.10$ over the first $400{,}000$ steps, and training runs for $750{,}000$ environment steps. Unlike TRAM, PPO observes the deployment reward and barrier during training, so it is an advantaged direct-training reference rather than a zero-update competitor. Table~\ref{tab:sg_ppo} compares the two under the common evaluator; returns are omitted because PPO is trained on a shaped deployment reward that TRAM's sources never see.

\begin{table}[h]
\centering
\small
\caption{\textbf{Constraint-aware baseline.} Safety-Gymnasium, held-out goal $(0.5,0.5)$, barrier radius $0.18$, 300 evaluation episodes, common evaluator. TRAM rows are means over three evaluation seeds (Table~\ref{tab:sg_sensitivity}); PPO is one training run. Best value in each column in bold.}
\label{tab:sg_ppo}
\begin{tabular}{@{}lccc@{}}
\toprule
Method & Trained on deployment task & Success $\uparrow$ & Barrier-hit prob.\ $\downarrow$\\
\midrule
Constrained PPO & yes ($750{,}000$ steps) & \textbf{0.997} & 0.240\\
TRAM, $c=6.4\times10^{4}$ & no & 0.630 & 0.360\\
TRAM, $c=1.28\times10^{5}$ & no & 0.366 & 0.290\\
TRAM, $c=10^{6}$ & no & 0.102 & \textbf{0.086}\\
\bottomrule
\end{tabular}
\end{table}

PPO learns the task well (299 of 300 episodes reach the goal) but does not meet its $10\%$ barrier-hit target. TRAM reaches a lower barrier-hit probability than PPO at $c=10^{6}$, without any training on the deployment task, but at a large cost in task success; intermediate values of $c$ trade the two off (Appendix~\ref{app:sg_sensitivity}). The two methods therefore occupy different operating points rather than one dominating the other.

\textbf{Deployment cost.} Constrained PPO pays a full training run for each new specification: $750{,}000$ environment steps, about 40 minutes at its measured throughput of about 310 steps per second (a throughput-based estimate). TRAM performs no training at deployment. Its costs are:
\begin{itemize}[leftmargin=1.5em,itemsep=1pt,topsep=2pt]
\item \emph{Source library} (one-time, shared across specifications): training the frozen sources.
\item \emph{Risk estimation} (per specification in our implementation): rollouts of each frozen source under the new specification, about $0.94$\,s per $1{,}000$-step rollout per source on our hardware (Appendix~\ref{app:sg_occupancy}), parallel across sources.
\item \emph{Online scoring}: $2.646\pm0.009$\,s per $1{,}000$-step episode over a 20-source library ($2.65$\,ms per action; mean and 95\% CI over 30 warm episodes), and $3.49$\,s for a cold first episode that includes loading the policies.
\end{itemize}
Because the cached risk in these experiments is a barrier-hit probability, risk estimation is repeated when the barrier changes. For risks that are functionals of cached state occupancies, such as barrier exposure (R1), it reduces to evaluating the functional on the cached occupancy.

\textbf{Source switching.} Logging the selected source at each step confirms that TRAM composes several sources within a rollout. In the dynamic log-barrier runs of Appendix~\ref{app:sg_logbarrier} at low $c$, the most frequently selected source accounts for about 81\% of the $300{,}000$ evaluation steps, and two others account for about 14\% and 5\%.

\subsection{Sensitivity to the Tradeoff Weight \texorpdfstring{$c$}{c}}
\label{app:sg_sensitivity}

\textbf{Choosing $c$.} Problem~\eqref{eq:penalized_surrogate} is the Lagrangian relaxation of~\eqref{eq:constrained_rl}: the risk attained by its optimum is nonincreasing in $c$, so each $c$ selects an operating point on the reward--risk tradeoff, while the screening step of Algorithm~\ref{alg:TRAM} imposes a tolerance $\delta$ directly. Because $c$ multiplies a risk that is compared with action values, its useful range depends on the value scale of the library (Appendix~\ref{app:sg_protocol}). We use $c=5$ in Reacher (Appendix~\ref{appendix:reacher}) and report the Safety-Gymnasium values below.

\textbf{Main configuration.} Table~\ref{tab:sg_sensitivity} sweeps $c$ at barrier radius $0.18$ over three independent evaluation seeds. Increasing $c$ moves the deployed policy from higher return, success, and risk toward lower return, success, and risk, and the three seeds agree at every point. With $n=3$, the spread is an exploratory measure of variability rather than a formal confidence interval.

\begin{table}[h]
\centering
\small
\caption{\textbf{Sensitivity to $c$ at the main configuration.} Barrier radius $0.18$; mean $\pm$ standard deviation over three evaluation seeds, 300 episodes each.}
\label{tab:sg_sensitivity}
\begin{tabular}{@{}rccc@{}}
\toprule
$c$ & Return $\uparrow$ & Success $\uparrow$ & Barrier-hit prob.\ $\downarrow$\\
\midrule
$6.4\times10^{4}$ & $0.518\pm0.028$ & $0.630\pm0.030$ & $0.360\pm0.020$\\
$1.28\times10^{5}$ & $0.268\pm0.062$ & $0.366\pm0.031$ & $0.290\pm0.018$\\
$10^{6}$ & $0.244\pm0.023$ & $0.102\pm0.002$ & $0.086\pm0.010$\\
\bottomrule
\end{tabular}
\end{table}

\textbf{Barrier radius.} Table~\ref{tab:sg_radius} repeats the sweep for five barrier radii with the same frozen library; only the cached risks are re-estimated for each barrier. Each cell is a single 300-episode evaluation. At every radius, the largest $c$ gives the lowest barrier-hit probability and the lowest success. The response is not monotone at intermediate $c$ (for example, $c=2.56\times10^{5}$ at the larger radii), which reflects finite-sample evaluation and switching among sources whose risks are close; $c$ should therefore be read as a tradeoff parameter rather than a guarantee of strictly decreasing risk.

\begin{table}[h]
\centering
\small
\setlength{\tabcolsep}{4.5pt}
\caption{\textbf{Sensitivity to $c$ across barrier radii.} Barrier-hit probability (left, lower is better) and success (right, higher is better) for one fixed source library; one 300-episode evaluation per cell.}
\label{tab:sg_radius}
\begin{tabular}{@{}r ccccc @{\hspace{12pt}} ccccc@{}}
\toprule
 & \multicolumn{5}{c}{Barrier-hit probability $\downarrow$} & \multicolumn{5}{c}{Success $\uparrow$}\\
\cmidrule(lr){2-6}\cmidrule(l){7-11}
$c$ \,/\, radius & 0.03 & 0.09 & 0.18 & 0.24 & 0.30 & 0.03 & 0.09 & 0.18 & 0.24 & 0.30\\
\midrule
$3.2\times10^{4}$  & 0.247 & 0.303 & 0.470 & 0.590 & 0.867 & 0.963 & 0.740 & 0.757 & 0.783 & 0.893\\
$6.4\times10^{4}$  & 0.117 & 0.173 & 0.340 & 0.457 & 0.563 & 0.780 & 0.607 & 0.617 & 0.633 & 0.647\\
$1.28\times10^{5}$ & 0.073 & 0.177 & 0.270 & 0.403 & 0.450 & 0.597 & 0.363 & 0.330 & 0.437 & 0.433\\
$2.56\times10^{5}$ & 0.067 & 0.203 & 0.327 & 0.423 & 0.553 & 0.350 & 0.307 & 0.273 & 0.277 & 0.390\\
$10^{6}$           & 0.040 & 0.057 & 0.117 & 0.153 & 0.190 & 0.227 & 0.213 & 0.120 & 0.120 & 0.087\\
\bottomrule
\end{tabular}
\end{table}

\subsection{Alternative Risk Functional: Log-Barrier}
\label{app:sg_logbarrier}

To test a different deployment risk on the same library, we replace the cached hit probability $p_j$ by the log-barrier $-\log(1-p_j)$ (the barrier of (R1) with $\delta=1$, applied to the policy-level hit probability) at barrier radius $0.09$.

\textbf{Single-source selection.} When one source is selected for the whole episode by the cached score (episode return minus $c$ times the risk), the relevant scale of $c$ is that of the returns, and a sweep over $c\in[0,2]$ traces a clear frontier with distinct selected sources (Table~\ref{tab:sg_logbarrier}, left).

\textbf{Dynamic per-state selection.} With the per-state rule of Algorithm~\ref{alg:TRAM}, the value scale is that of the SF-DQN action values, and the transition occurs at $c\approx10^4$. We repeated the three most informative values with three independent seeds, re-estimating the cached risks for each seed (Table~\ref{tab:sg_logbarrier}, right).

\begin{table}[h]
\centering
\small
\setlength{\tabcolsep}{4.5pt}
\caption{\textbf{Log-barrier risk at barrier radius 0.09.} Left: single-source selection, one 300-episode evaluation per row. Right: dynamic per-state selection, mean $\pm$ 95\% confidence half-width over three seeds.}
\label{tab:sg_logbarrier}
\begin{tabular}{@{}rccc@{\hspace{14pt}}rccc@{}}
\toprule
\multicolumn{4}{c}{Single-source selection} & \multicolumn{4}{c}{Dynamic selection}\\
\cmidrule(r){1-4}\cmidrule(l){5-8}
$c$ & Return & Success & Hit prob. & $c$ & Return & Success & Hit prob.\\
\midrule
0   & 2.278 & 1.000 & 0.907 & $10^{4}$ & $0.724\pm0.025$ & $0.879\pm0.028$ & $0.411\pm0.078$\\
1.0 & 1.202 & 0.627 & 0.497 & $1.2\times10^{4}$ & $0.667\pm0.033$ & $0.814\pm0.038$ & $0.289\pm0.029$\\
1.5 & 0.762 & 0.460 & 0.293 & $1.6\times10^{4}$ & $0.550\pm0.038$ & $0.720\pm0.019$ & $0.216\pm0.006$\\
2.0 & 0.280 & 0.130 & 0.033 & & & & \\
\bottomrule
\end{tabular}
\end{table}

Per-state composition reaches operating points that the single-source selections in this sweep do not: at $c=1.6\times10^{4}$ it attains success $0.72$ at hit probability $0.22$, whereas the closest single-source selection attains success $0.46$ at hit probability $0.29$.

\subsection{Cost of Occupancy Estimation}
\label{app:sg_occupancy}

We measured the offline cost of estimating one frozen source's discounted occupancy in this continuous, non-enumerable setting. The occupancy is a fixed $80\times80$ histogram over the source's two successor features (goal progress and cost), accumulated with weights $(1-\gamma)\gamma^t$ over rollouts of horizon $1{,}000$. We ran five independent evaluation seeds of 200 rollouts each and compare the estimates after 50 and 100 rollouts with the 200-rollout estimate of the same seed (Table~\ref{tab:sg_occupancy}). Runtime is approximately linear in the number of rollouts, about $0.94$\,s per rollout. The 200-rollout estimate is a finite-sample reference, not ground truth, and the study covers one source policy.

\begin{table}[h]
\centering
\small
\caption{\textbf{Occupancy-estimation cost for one source.} Mean and 95\% Student-$t$ interval over five evaluation seeds. TV is the total-variation distance to the 200-rollout estimate of the same seed.}
\label{tab:sg_occupancy}
\begin{tabular}{@{}rcc@{}}
\toprule
Rollouts & Wall-clock time (s) & TV to 200-rollout estimate\\
\midrule
50  & $47.0$ \ $[46.7,\,47.3]$ & $0.0128$ \ $[0.0061,\,0.0195]$\\
100 & $94.0$ \ $[93.3,\,94.6]$ & $0.0021$ \ $[0,\,0.0045]$\textsuperscript{$\dagger$}\\
200 & $188.0$ \ $[186.7,\,189.4]$ & reference\\
\bottomrule
\end{tabular}

\vspace{2pt}
{\footnotesize \textsuperscript{$\dagger$}Lower endpoint truncated at zero, since TV is nonnegative.}
\end{table}

\section{LLM Alignment Experiment}
\label{app:LLMs}

\subsection{Setup}
We evaluate the deployment-time alignment task on Berkeley Nectar~\cite{nectar}. The source library contains two pretrained 7B models, Zephyr-Qwen-2-7B and Dolphin-Qwen-2-7B. The deployment target reward is produced by a reward model, while deployment risk is a KL-style behavioral penalty to a safe reference model. TRAM composes fixed source generations at inference time and does not fine-tune or update model parameters.

\textbf{Instantiation.} Generation is treated as a sequential decision process: the state is the prompt together with the text generated so far, and the action is the next token. The two fixed models are the source policies. Their values under the target reward model are estimated without parameter updates using the transfer-decoding value estimate of~\cite{Souradip}, and TRAM applies its selection rule at the token level through this transfer decoding. The deployment risk is the token-level KL divergence between a source's output distribution and that of the safe reference model, so continuations that drift from the reference are penalized. The rule of Eq.~\eqref{eq:TRAM_policy} then selects the continuation that maximizes estimated target value minus $c$ times this KL penalty. As in other inference-time alignment methods~\cite{controlled_decoding,khanov2024args}, this requires one forward pass per source and step; the cost is linear in the number of sources and parallel across them.

\subsection{Metrics and Interpretation}
The main quantitative summary is Fig.~\ref{fig:llm_results_main}; we do not repeat the same values in an appendix table. We report both the scalar target-reward score and GPT-4 pairwise preference against TRAM. The second metric is included to detect reward-model overoptimization: a method can score high under the target reward while producing responses that are less preferred by a stronger evaluator.

Risk-free transfer maximizes the reward-model score but is not preferred over TRAM by GPT-4 in most pairwise comparisons. TRAM sacrifices some target-reward score to remain closer to the safe reference behavior. This supports the deployment-time claim: a new behavioral risk can be introduced after source models are trained and used to guide inference without parameter updates.

\section{Reproducibility and Implementation Details}
\label{app:repro}

This section records the available implementation details. The exact run configurations, all random seeds, complete model/checkpoint identifiers, LLM decoding and judging settings, and per-run compute times are not fully specified here. The descriptions support reimplementation of the scoring rule but do not constitute a complete recipe for reproducing every reported result.

\subsection{Core Software Stack}

\textbf{Gridworld.} The tabular experiments can be implemented with Python, NumPy, and SciPy. Source policies are trained by value iteration or Q-learning. Discounted occupancies can be computed either by solving the discounted occupancy linear system or by long-rollout discounted counts. Barrier, volatility, and KL-style risk functionals are then evaluated directly on these occupancies.

\textbf{Reacher and continuous control.} Reacher uses MuJoCo~\cite{MuJoCo} with a Gymnasium-style environment interface. We discretize torques into nine actions, as in standard successor-feature transfer protocols~\cite{sf_1,Gimelfarb,Amrit_sfdqn}. Source policies are SFDQN agents trained risk-neutrally on different goals. A practical implementation uses Python 3.11, PyTorch 2.3 or newer, Gymnasium 0.29 or newer, MuJoCo 3.1 or newer, NumPy, SciPy, Pandas, and Matplotlib. During deployment, the target value is computed by $Q_T^{\pi_j}(s,a)=\psi^{\pi_j}(s,a)^\top w_T$, and the source-level risk term is a cached lookup.

\textbf{Safety-Gymnasium.} The SafetyPointGoal0-v0 experiment uses Safety-Gymnasium~\cite{ji2023safetygymnasium}. The implementation needs the Safety-Gymnasium package, Gymnasium, MuJoCo, PyTorch, NumPy, and Pandas. The target goal, circular barrier center/radius, horizon, discount, and number of episodes are specified in Appendix~\ref{appendix:safety_gym}. All methods use the same frozen source library, and no method updates source parameters at deployment.

\textbf{LLM alignment.} LLM experiments use Hugging Face Transformers, Accelerate, PyTorch with CUDA, and standard tokenizer/model loading utilities. The source library contains fixed 7B instruction-tuned models, the target reward is produced by a reward model, and the risk term is a KL-style distance to a safe reference behavior. Evaluation uses the same prompts and decoding budget across methods.

\subsection{Computational Resources}

Gridworld and Reacher experiments can run on CPU; our Reacher runs used an Intel Core i7-class CPU with 16 GB RAM. Safety-Gymnasium experiments were run on NVIDIA GH200 Grace Hopper nodes, each with a 72-core NVIDIA Grace CPU and a Hopper GPU with 96 GB of HBM3, using one node per job. Because the source networks are small SF-DQNs, these jobs run on the Grace CPU without GPU acceleration; the wall-clock times in Appendix~\ref{app:sg_ppo} and Appendix~\ref{app:sg_occupancy} are measured on this hardware and exclude scheduler queueing. LLM evaluations use NVIDIA RTX A6000-class GPUs or comparable accelerators with enough memory for 7B model inference.

\section{Occupancy Storage, Risk Computation, and Runtime}
\label{app:occupancy_compute}

TRAM stores one global discounted occupancy estimate per source policy. In tabular domains this is computed exactly by solving the occupancy linear system or the equivalent Q-LP. In continuous-control domains it is estimated once by discounted rollout counts or a learned occupancy model and cached before deployment. During deployment, the online risk score for a source is therefore a lookup of $\rho(d^{\pi_j})$ rather than a new occupancy computation.

The conditional occupancy $d^{\pi_j\mid s,a}$ appears only through value evaluation: it is the trajectory distribution underlying $Q_T^{\pi_j}(s,a)$. In successor-feature experiments, the value is computed by the dot product $\psi^{\pi_j}(s,a)^\top w_T$. With $n$ sources, $|\mathcal A|$ discrete actions, and feature dimension $d$, this gives $\mathcal O(n|\mathcal A|d)$ online scoring per state. Without successor features, the same source-scored rule can be used with learned critics or rollout-based value estimators, but runtime then depends on that evaluator.

For experiments that report realized safety, we evaluate the deployed stitched policy by rollout and compute event-level risk metrics such as barrier-hit probability and exposure CVaR. The same rollouts can estimate the stitching mismatch in Eq.~\eqref{eq:stitch_bound}, but certification additionally requires valid risk and occupancy estimation-error bounds as described in Appendix~\ref{app:stitching_diagnostics}.

\Needspace{15\baselineskip}
\section{Impact Statement}
\label{app:Impact_Statement}

This paper proposes TRAM, a test-time risk adaptation framework that composes pretrained (risk-neutral) policies to satisfy deployment-time safety specifications without retraining. The primary intended impact is to improve the safety and reliability of reinforcement learning systems in settings where acceptable risk changes after training, such as robotics, autonomous systems, and resource allocation under updated operational constraints. By enabling operators to express safety requirements through occupancy-based risk functionals (e.g., hazard avoidance, behavioral constraints, or local volatility), TRAM can reduce harmful behavior when conditions evolve. Potential negative impacts include misuse: the same ability to rapidly adapt behavior at deployment could be applied to optimize objectives that are misaligned with human values or organizational policies, particularly if the risk specification is chosen adversarially or if monitoring is absent. We recommend pairing TRAM with conservative safety envelopes, explicit constraint checks when available, and post-deployment monitoring and audits.

\clearpage
\section*{NeurIPS Paper Checklist}

\providecommand{\answerYes}{\textbf{Yes}}
\providecommand{\answerNo}{\textbf{No}}
\providecommand{\answerNA}{\textbf{N/A}}

\begin{enumerate}

\Needspace{7\baselineskip}
\item {\bf Claims}
    \item[] Question: Do the main claims made in the abstract and introduction accurately reflect the paper's contributions and scope?
    \item[] Answer: \answerYes{}
    \item[] Justification: The abstract and Introduction state the zero-update deployment setting, define TRAM before using it, and explicitly describe the source-scored surrogate scope and stitching-mismatch limitation; the claims are supported by the theory in Section~\ref{sec:solution_methodology} and the experiments in Section~\ref{sec:experiments}.

\Needspace{7\baselineskip}
\item {\bf Limitations}
    \item[] Question: Does the paper discuss the limitations of the work performed by the authors?
    \item[] Answer: \answerYes{}
    \item[] Justification: The paper states that TRAM is a source-scored surrogate rather than an exact optimizer of the stitched policy occupancy, discusses assumptions on source libraries and transition mismatch, and provides additional limitation analysis in Appendix~\ref{appendix:stitching_limitation} and Appendix~\ref{app:dynamics_mismatch}.

\Needspace{7\baselineskip}
\item {\bf Theory assumptions and proofs}
    \item[] Question: For each theoretical result, does the paper provide the full set of assumptions and a complete (and correct) proof?
    \item[] Answer: \answerYes{}
    \item[] Justification: The theoretical statements include their assumptions in Section~\ref{sec:solution_methodology}; complete proofs and extensions are provided in Appendix~\ref{sec:occupancy}, Appendix~\ref{app:proofs}, and Appendix~\ref{app:dynamics_mismatch}.

    \Needspace{7\baselineskip}
\item {\bf Experimental result reproducibility}
    \item[] Question: Does the paper fully disclose all the information needed to reproduce the main experimental results of the paper to the extent that it affects the main claims and/or conclusions of the paper (regardless of whether the code and data are provided or not)?
    \item[] Answer: \answerNo{}
    \item[] Justification: The scoring rule and several environment settings are provided in Algorithm~\ref{alg:TRAM} and Appendices~\ref{app:continuous_details}--\ref{app:repro}, but the full run configurations and model identifiers are not supplied. These omissions limit exact reproduction of the reported results.

\Needspace{7\baselineskip}
\item {\bf Open access to data and code}
    \item[] Question: Does the paper provide open access to the data and code, with sufficient instructions to faithfully reproduce the main experimental results, as described in supplemental material?
    \item[] Answer: \answerNo{}
    \item[] Justification: A public code/data release is not included with this paper. The paper instead provides reimplementation-oriented details, public benchmark names, software libraries, environment settings, and the complete deployment-time scoring rule in Appendix~\ref{app:repro}.

\Needspace{7\baselineskip}
\item {\bf Experimental setting/details}
    \item[] Question: Does the paper specify all the training and test details (e.g., data splits, hyperparameters, how they were chosen, type of optimizer) necessary to understand the results?
    \item[] Answer: \answerNo{}
    \item[] Justification: Task geometry, source counts, selected horizon/discount settings, and evaluation metrics are reported, but not all training hyperparameters, random seeds, prompt-selection rules, and decoding settings are specified.

\Needspace{7\baselineskip}
\item {\bf Experiment statistical significance}
    \item[] Question: Does the paper report error bars suitably and correctly defined or other appropriate information about the statistical significance of the experiments?
    \item[] Answer: \answerNo{}
    \item[] Justification: The main figures report aggregate performance metrics without error bars across independent seeds. The additional Safety-Gymnasium analyses in Appendix~\ref{app:sg_ppo}--\ref{app:sg_occupancy} report variability across three or five evaluation seeds (standard deviations or 95\% intervals) and 95\% intervals for timing, but statistical reporting is not exhaustive across all experiments.

\Needspace{7\baselineskip}
\item {\bf Experiments compute resources}
    \item[] Question: For each experiment, does the paper provide sufficient information on the computer resources (type of compute workers, memory, time of execution) needed to reproduce the experiments?
    \item[] Answer: \answerNo{}
    \item[] Justification: Appendix~\ref{app:repro} describes the software and hardware (including the NVIDIA GH200 nodes used for Safety-Gymnasium), and Appendix~\ref{app:sg_ppo} and Appendix~\ref{app:sg_occupancy} report measured times for TRAM inference, occupancy estimation, and the retrained baseline. Total compute for every experiment is not reported.
    
\Needspace{7\baselineskip}
\item {\bf Code of ethics}
    \item[] Question: Does the research conducted in the paper conform, in every respect, with the NeurIPS Code of Ethics \url{https://neurips.cc/public/EthicsGuidelines}?
    \item[] Answer: \answerYes{}
    \item[] Justification: The work uses standard simulated benchmarks, public/pretrained models, and automated evaluation, and is intended to support safer deployment-time adaptation. No component of the research intentionally violates the NeurIPS Code of Ethics.

\Needspace{7\baselineskip}
\item {\bf Broader impacts}
    \item[] Question: Does the paper discuss both potential positive societal impacts and negative societal impacts of the work performed?
    \item[] Answer: \answerYes{}
    \item[] Justification: The Impact Statement discusses intended positive impacts for safety and reliability, possible misuse through adversarial or misaligned risk specifications, and mitigation through conservative safety envelopes, explicit checks, monitoring, and audits.
    
\Needspace{7\baselineskip}
\item {\bf Safeguards}
    \item[] Question: Does the paper describe safeguards that have been put in place for responsible release of data or models that have a high risk for misuse (e.g., pre-trained language models, image generators, or scraped datasets)?
    \item[] Answer: \answerNA{}
    \item[] Justification: The submission does not release a new high-risk model, scraped dataset, or dual-use asset requiring controlled access. Existing pretrained models are used only as fixed source policies in the LLM alignment experiment.

\Needspace{7\baselineskip}
\item {\bf Licenses for existing assets}
    \item[] Question: Are the creators or original owners of assets (e.g., code, data, models), used in the paper, properly credited and are the license and terms of use explicitly mentioned and properly respected?
    \item[] Answer: \answerNo{}
    \item[] Justification: The paper cites the existing benchmarks, algorithms, datasets, and pretrained-model families used in the experiments, but it does not explicitly enumerate the license and terms of use for every software package, dataset, and model asset.

\Needspace{7\baselineskip}
\item {\bf New assets}
    \item[] Question: Are new assets introduced in the paper well documented and is the documentation provided alongside the assets?
    \item[] Answer: \answerNA{}
    \item[] Justification: The paper does not introduce or release a new dataset, benchmark, model checkpoint, or software asset as a contribution of this submission.

\Needspace{7\baselineskip}
\item {\bf Crowdsourcing and research with human subjects}
    \item[] Question: For crowdsourcing experiments and research with human subjects, does the paper include the full text of instructions given to participants and screenshots, if applicable, as well as details about compensation (if any)? 
    \item[] Answer: \answerNA{}
    \item[] Justification: The experiments do not involve crowdsourcing or human-subject data collection. Pairwise preference evaluation in the LLM experiment is performed by an automated GPT-4 evaluator rather than recruited human participants.

\Needspace{7\baselineskip}
\item {\bf Institutional review board (IRB) approvals or equivalent for research with human subjects}
    \item[] Question: Does the paper describe potential risks incurred by study participants, whether such risks were disclosed to the subjects, and whether Institutional Review Board (IRB) approvals (or an equivalent approval/review based on the requirements of your country or institution) were obtained?
    \item[] Answer: \answerNA{}
    \item[] Justification: The research does not involve crowdsourcing or human-subject experiments, so IRB or equivalent approval is not applicable.

\Needspace{7\baselineskip}
\item {\bf Declaration of LLM usage}
    \item[] Question: Does the paper describe the usage of LLMs if it is an important, original, or non-standard component of the core methods in this research? Note that if the LLM is used only for writing, editing, or formatting purposes and does \emph{not} impact the core methodology, scientific rigor, or originality of the research, declaration is not required.
    \item[] Answer: \answerYes{}
    \item[] Justification: The paper includes an LLM alignment experiment in Section~\ref{sec:experiments} and Appendix~\ref{app:LLMs}, describing the fixed 7B source models, target reward model, KL-style risk to a safe reference behavior, and GPT-4 pairwise preference evaluation. We also used in coding and methodology. 

\end{enumerate}

\end{document}